\newcommand{\Real}{\mathbb R}

\newcommand{\Eq}[1]{\begin{equation}#1\end{equation}}
\newcommand{\Vc}[1]{\mbox{\boldmath$#1$}}

\newcommand{\vx}{\Vc{x}}
\newcommand{\vX}{\Vc{X}}

\newcommand{\vY}{\Vc{Y}}

\newcommand{\mX}{\mathcal{X}}
\newcommand{\mI}{\mathcal{I}}

\newcommand{\mP}{\mathcal{P}}

\DeclareMathAlphabet{\mathsfsl}{OT1}{cmss}{m}{sl}

\documentclass[12pt,journal,draftcls,onecolumn]{IEEEtran}

\usepackage{pifont,calc,amsmath,amsfonts,amsbsy,amssymb,tabularx,float,times}
\usepackage{cite,url,graphicx,subfigure,epsfig,algorithm,algorithmic,multirow}

\begin{document}

\title{Information Preserving Component Analysis: Data Projections for Flow Cytometry Analysis}%
\author{Kevin M. Carter$^1$\thanks{{\bf Acknowledgement}: This work is
partially funded by the National Science Foundation, grant No.
CCR-0325571.}, Raviv Raich$^2$, William G. Finn$^3$, and Alfred O. Hero III$^1$\\
$^{1}$ Department of EECS,    University of Michigan,    Ann Arbor,
MI 48109\\
$^{2}$ School of EECS, Oregon State University, Corvallis, OR 97331\\
$^{3}$ Department of Pathology,    University of Michigan,    Ann Arbor,
MI 48109\\
{\normalsize \tt  \{kmcarter,wgfinn,hero\}@umich.edu}, {\normalsize \tt
raich@eecs.oregonstate.edu}}
\maketitle

\begin{abstract}
Flow cytometry is often used to characterize the malignant cells in leukemia and lymphoma patients, traced to the level of the individual cell. Typically, flow cytometric data analysis is performed through a series of 2-dimensional projections onto the axes of the data set. Through the years, clinicians have determined combinations of different fluorescent markers which generate relatively known expression patterns for specific subtypes of leukemia and lymphoma -- cancers of the hematopoietic system. By only viewing a series of 2-dimensional projections, the high-dimensional nature of the data is rarely exploited. In this paper we present a means of determining a low-dimensional projection which maintains the high-dimensional relationships (i.e.~information) between differing oncological data sets. By using machine learning techniques, we allow clinicians to visualize data in a low dimension defined by a linear combination of all of the available markers, rather than just 2 at a time. This provides an aid in diagnosing similar forms of cancer, as well as a means for variable selection in exploratory flow cytometric research. We refer to our method as Information Preserving Component Analysis (IPCA).

\end{abstract}

\begin{keywords}
Flow cytometry, statistical manifold, information geometry, multivariate data analysis, dimensionality reduction, clustering
\end{keywords}
\section{Introduction}
Clinical flow cytometric data analysis usually involves the interpretation of data culled from sets (i.e.~cancerous blood samples) which contain the simultaneous analysis of several measurements. This high-dimensional data set allows for the expression of different fluorescent markers, traced to the level of the single blood cell. Typically, diagnosis is determined by analyzing individual 2-dimensional scatter plots of the data, in which each point represents a unique blood cell and the axes signify the expression of different biomarkers. By viewing a series of these histograms, a clinician is able to determine a diagnosis for the patient through clinical experience of the manner in which certain leukemias and lymphomas express certain markers.

Given that the standard method of cytometric analysis involves projections onto the axes of the data (i.e.~visualizing the scatter plot of a data set with respect to 2 specified markers), the multi-dimensional nature of the data is not fully exploited. As such, typical flow cytometric analysis is comparable to hierarchical clustering methods, in which data is segmented on an axis-by-axis basis. Marker combinations have been determined through years of clinical experience, leading to relative confidence in analysis given certain axes projections. These projection methods, however, contain the underlying assumption that marker combinations are independent of each other, and do not utilize the dependencies which may exist within the data. Ideally, clinicians would like to analyze the full-dimensional data, but this cannot be visualized outside of 3-dimensions.

There have been previous attempts at using machine learning to aid in flow cytometry diagnosis. Some have focused on clustering in the high-dimensional space \cite{Zeng&EtAl:JBI07,Zamir&EtEal:BJH05}, while others have utilized information geometry to identify differences in sample subsets and between data sets \cite{Roederer&Hardy:Cyt01-1,Roederer&Hardy:Cyt01-2}. These methods have not satisfied the problem because they do not significantly approach the aspect of visualization for `human in the loop' diagnosis, and the ones that do \cite{Mann&Popp:Cyt84,Mann:Cyt87} only apply dimensionality reduction to a single set at a time. The most relevant work, compared to what we are about to present, is that which we have recently presented  \cite{Finn&Carter:CytB08} where we utilized information geometry to simultaneously embed each patient data set into the same low-dimensional space, representing each patient as a single vector. The current task differs in that we do not wish to reduce each set to a single point for comparative analysis, but to use dimensionality reduction as a means to individually study the distributions of each patient. As such, we aim to reduce the dimension of each patient data set while maintaining the number of data points (i.e.~cells).


With input from the Department of Pathology at the University of Michigan, we have determined that the ideal form of dimensionality reduction for flow cytometric visualization would contain several properties. The data needs to be preserved without scaling or skewing, as this is most similar to the current methods in practice (i.e.~axes projections). Hence, the ideal projection should be orthonormal. Secondly, the methods should be unsupervised, relying solely on the geometry of the data. This requirement is straight forward as the dimensionality reduction would be an aid for diagnosis, so no labels would be available. As such, common supervised methods geared towards dimensionality reduction for classification tasks (e.g.~LDA methods \cite{Friedman:JASA89,Mika:NN99}) are not applicable towards this problem.

Clinicians would also like to work in a low-dimensional space similar to what they have grown accustomed to through years of experience. Once determined, the subspace should be consistent, and should not change when processing new data. Therefore non-linear methods of dimensionality reduction such as \cite{Tenenbaum&etal:Science00,Roweis&Saul:Science00} are not ideal for this task. Adding new data to non-linear methods forces a re-computation of the subspace, which may be noticeably different than previous spaces (e.g.~scaled or rotated differently). This has been approached with out-of-sample extension methods \cite{Bengio&EtAl:NIPS04}, but it is still a relatively open problem. Finally, the projection space needs to preserve the relationship between data sets; patients in the same disease class should show similar expressions in the low-dimensional space, while differing disease classes should be distinct from one another. This requirement leads directly to a projection method which maintains the similarity between multiple data sets, rather than preserving similarities between the elements of a single set.

Given the desired properties, we present a method of dimensionality reduction -- which we refer to as \emph{Information Preserving Component Analysis (IPCA)} -- that preserves the Fisher information between data sets. We have shown in previous work \cite{Carter:TSP08, Carter&Raich:ICASSP08} that the Fisher information distance is the appropriate means for determining the similarity between non-Euclidean data. This is the case for flow cytometry data, as certain channels may represent light scatter angles, while other channels correspond to the expression of a specific fluorescent marker. Hence, there is no straight-forward Euclidean representation of the data.

IPCA operates in the space of linear and unsupervised dimensionality reduction methods, such as Principal Components Analysis (PCA), Projection Pursuit (PP) \cite{Friedman&Tukey:TC74} and Independent Component Analysis (ICA) \cite{Hyvarinen:ICA01}. By preserving the Fisher information distance between sets, IPCA ensures that the low-dimensional representation maintains the similarities between data sets which are contained in the full-dimensional data, minimizing the loss of information. This low-dimensional representation is a linear combination of the various markers, enabling clinicians to visualize all of the data simultaneously, rather than the current process of axes projections, which only relays information in relation to two markers at a time. Additionally, analysis of the loading vectors within the IPCA projection matrix offers a form of variable selection, which relays information describing which marker combinations yield the most information. This has the significant benefit of allowing for exploratory data analysis.

This paper proceeds as follows: In Section \ref{s:background} we give a background of flow cytometry and the typical clinical analysis process, as well as a formulation of the problem we will attempt to solve. We present our methods for finding the IPCA projection in Section \ref{s:methods}. Simulation results for both synthetic and clinical cytometric data are illustrated in Section \ref{s:results}, followed by a discussion and areas for future work in Section \ref{s:conclusions}.

\section{Background}
\label{s:background}
Clinical flow cytometry is widely used in the diagnosis and management of malignant disorders of the blood, bone marrow, and lymph nodes (leukemia and lymphoma).  In its basic form, flow cytometry involves the transmission of a stream of cells through a laser light source, with characteristics of each cell determined by the nature of the light scattered by the cell through disruption of the laser light.  Application to leukemia and lymphoma diagnosis is usually in the form of flow cytometric immunophenotyping, whereby cells are labeled with antibodies to specific cellular antigens, and the presence of these antigens detected by light emitted from fluorescent molecules (of different ``colors'') conjugated to the target antibody.

Clinical grade flow cytometers typically assess the size and shape of cells through the detection of light scattered at two predetermined angles (forward angle light scatter, and side angle or orthogonal light scatter), and are also capable of simultaneously detecting the expression patterns of numerous cellular antigens in a single prepared cell suspension (``tube'').  The analysis of multiple tubes then allows for any number of antigen expression patterns to be assessed.  Although 8-color flow cytometry is possible with the latest generation of clinical grade analyzers, most clinical flow cytometry laboratories utilize 3 or 4 color approaches.


\begin{figure*}[t]
  \centerline{
  \subfigure[CD5 vs. CD19]{
    \includegraphics[scale=.36]{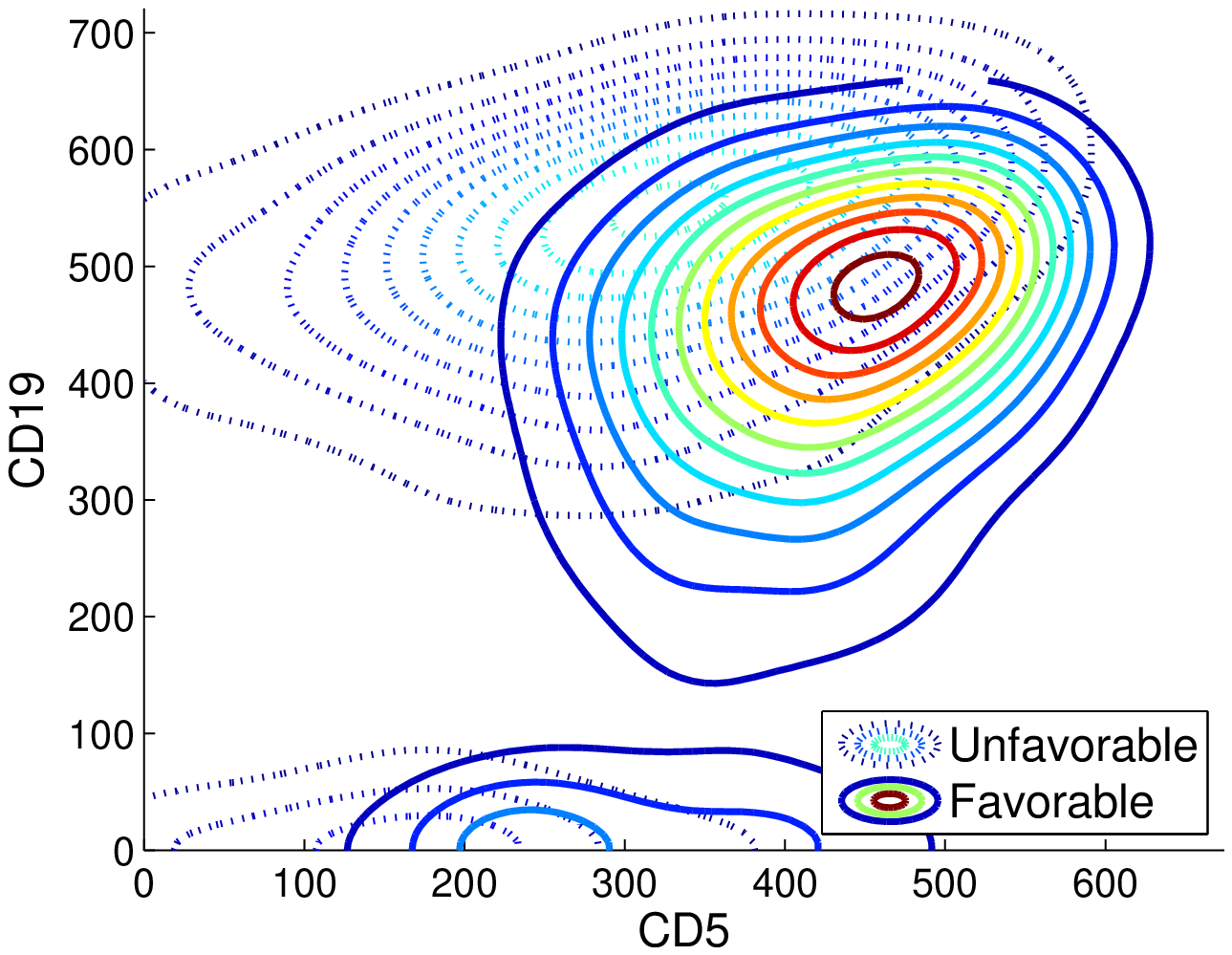}\\}
  \subfigure[CD5 vs. CD38]{
    \includegraphics[scale=.36]{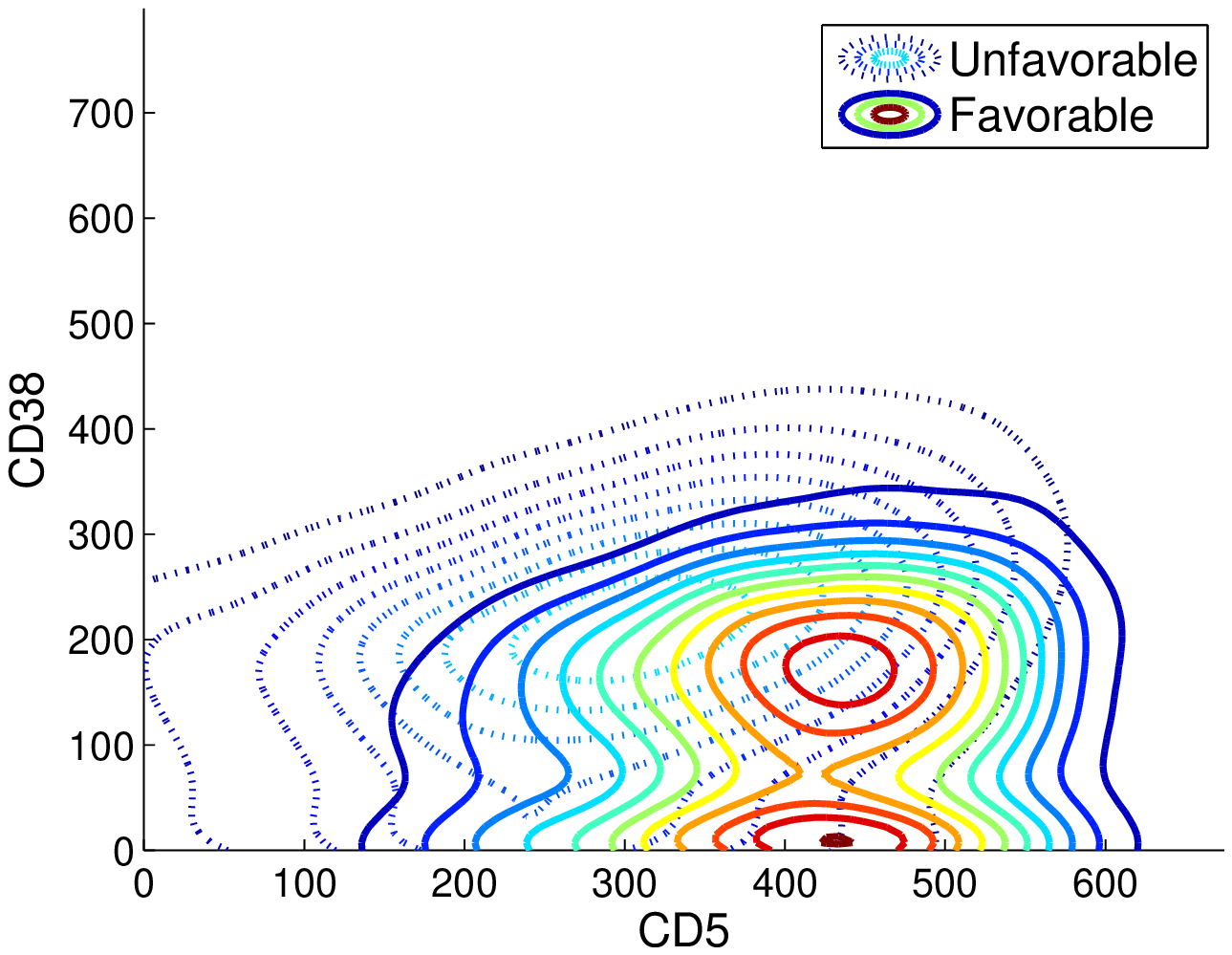}\\}
  \subfigure[CD38 vs. CD19]{
    \includegraphics[scale=.36]{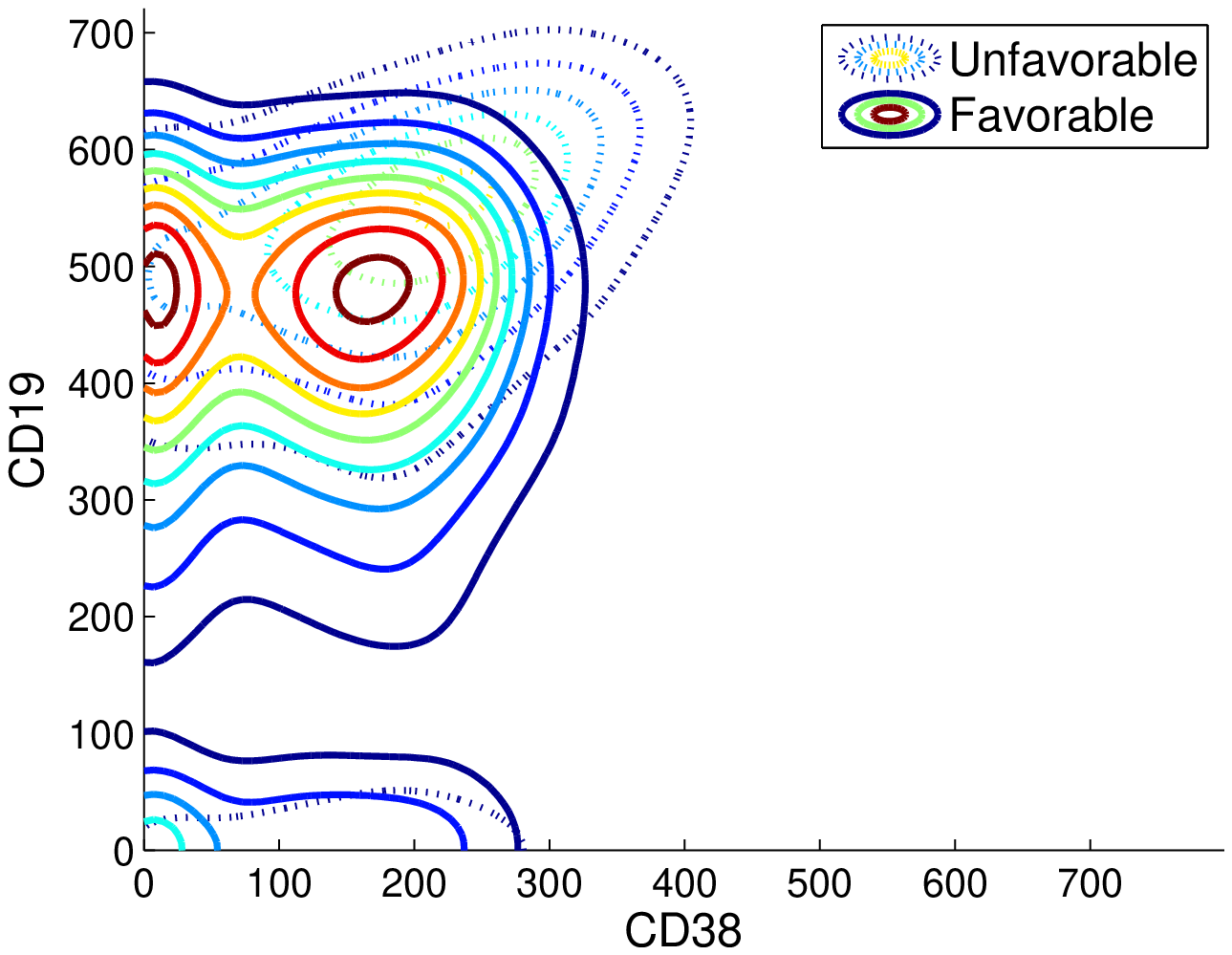}\\}
  }
  \caption{Typically, flow cytometric analysis is performed using multiple 2-dimensional projections onto the various marker combinations. This can lead to ambiguity and does not fully exploit the high-dimensional nature of the data. We illustrate this difficulty in distinguishing a patient with an unfavorable immunophenotype to that of a favorable patient, using their marginal PDFs over 3 of a possible 60 marker combinations from the 6-marker assay.}
  \label{f:cytometry_process}
\end{figure*}

In routine flow cytometric immunophenotyping, the expression patterns of each marker in a given tube can be traced to the level of the single cell, giving flow cytometry a uniquely spatial characteristic when compared to other immunophenotyping or proteomic analysis methods.  When measurements of forward and side angle light scatter characteristics are included, each cell analyzed via 4-color flow cytometry can be thought of as occupying a unique point in 6-dimensional space, with the dimensions of each point defined by the magnitude of expression of each antigen or light scatter characteristic.  Since all 6 dimensions cannot be projected simultaneously onto a single histogram, diagnosticians typically analyze a series of 2-dimensional histograms defined by any 2 of the 6 characteristics measured in a given tube (see Fig.~\ref{f:cytometry_process}).  Often one or more measured characteristics are used to restrict immunophenotypic analysis to a specific subset of cells in a process commonly known as \emph{gating}, which allows for limited exploitation of the dimensionality of the flow cytometry data set.

The use of each single measured characteristic as an axis on a 2-dimensional histogram is a convenient method for visualizing results and observing relationships between cell surface markers, but is equivalent to viewing a geometric shape head-on, and therefore does not necessarily take full advantage of the multidimensional nature of flow cytometry.  Just as it is possible to rotate an object in space to more effectively observe that object's characteristics, so too is it possible to ``rotate'' the 2-dimensional projection of a 6-dimensional flow cytometry analysis to optimally view the relationships among the 6 measured characteristics.
\subsection{Problem Formulation}

\begin{figure}[t]
  \center
    \includegraphics[scale=.55]{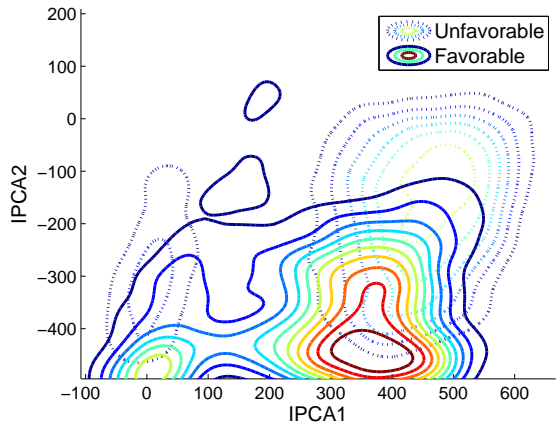}\\
  \caption{Projecting the same data in Fig.~\ref{f:cytometry_process} down to 2-dimensions using a linear combination of all available markers. It is a much easier task to discriminate these joint PDFs.}
  \label{f:ipca_dens}
\end{figure}

Given the critical importance of visualization in the task of flow cytometric diagnosis, we wish to find the low-dimensional projection which best preserves the relationships between patient data sets. Rather than viewing a series of axes projections determined by clinical experience as in Fig.~\ref{f:cytometry_process} (where we illustrate only 3 of the 60 possible axes projections of the 6-dimensional data set), a projection which is a linear combination of several biomarkers will allow a clinician to visualize all of the data in a single low-dimensional space, with minimal loss of information. An example is shown in Fig.~\ref{f:ipca_dens}, where it is easy to differentiate the patient with an unfavorable immunophenotype from that of a favorable patient\footnote{The data presented here is from patients with chronic lymphocytic leukemia, and is further explained in Section \ref{ss:cll}}.

Specifically, given a collection of flow cytometer outputs $\mX=\left\{\vX_1,\ldots,\vX_N\right\}$ in which each element of $\vX_i$ exists in $\Real^d$, we can define similarity between data sets $\vX_i$ and $\vX_j$ (e.g.~patients $i$ and $j$) with some metric as $D(\vX_i,\vX_j)$. Can we find a mapping \[A:\vX\rightarrow \vY\] in which the elements of $\vY$ exist in $\Real^m$, $m<d$ ($m=2$ or $3$ for visualization) such that \[D(\vX_i,\vX_j)=D(\vY_i,\vY_j), \; \forall \: i,j?\] Can we define this mapping as a linear projection $A\in\Real^{m\times d}$? Can we ensure that the projection minimally alters the data itself (i.e.~ensure $A$ is orthonormal)? Additionally, by analyzing the loadings in $A$, can we determine which biomarkers are best at differentiating between disease classes?

\section{Methods}
\label{s:methods}


In our previous work on Fisher Information Non-parametric Embedding (FINE) \cite{Carter:TSP08,Finn&Carter:CytB08}, we have shown that we can derive an information-based embedding for the purposes of flow cytometric analysis (See Appendix \ref{AP:FINE}). By viewing each patient as a probability density function (PDF) on a statistical manifold, we were able to embed that manifold into a low-dimensional Euclidean space, in which each patient is represented by a single point. This visualization allows for a diagnostician to view each patient in relation to other selected patients in a space where disease classes are well distinguished. The similarity between patients was determined by using an approximation of the Fisher information distance between PDFs parameterized by $\theta=[\theta^1,\ldots,\theta^n]$:
\Eq{ \label{Eq:FID_multi}
    D_F(\theta_1,\theta_2)=\min_{\theta:\theta(0)=\theta_1,\theta(1)=\theta_2}
        \int_{0}^{1}{\sqrt{\left(\frac{d\theta}{d\beta}\right)^T \mI(\theta) \left(\frac{d\theta}{d\beta}\right)} d\beta}
    ,}
where $\theta_1$ and $\theta_2$ are parameter values corresponding to the two PDFs and $\mI(\theta)$ is the Fisher information matrix whose elements are
\Eq{ \label{Equation:FIM}
\mI_{ij}=\int{f(X;\theta)\frac{\partial \log f(X;\theta)}{\partial\theta^i}\frac{\partial \log f(X;\theta)}{\partial\theta^j}\,dX}
.}

The Fisher information distance is the best way to characterize similarity between PDFs as it is an exact measure of the geodesic (i.e.~shortest path) between points along the manifold. While the Fisher information distance cannot be exactly computed without knowing the parameterization of the manifold, it may be approximated with metrics such as the Kullback-Leibler (KL) divergence, Hellinger distance, and R\'{e}nyi-alpha entropy \cite{Kass&Vos:97}. For our work, we focus on the KL-divergence, which is defined as
\Eq{\label{Equation:KL}
KL(p_1\|p_2)=\int{p_1(x)\log \frac{p_1(x)}{p_2(x)}}
 ,}
where $p_1$ and $p_2$ are PDFs of possibly unknown parameterization. It should be noted that the KL-divergence is not a distance metric, as it is not symmetric, $KL(p_1\|p_2)\neq KL(p_1\|p_2)$. To obtain this symmetry, we will define the KL-divergence as:
\Eq{ \label{Equation:DKL}
        D_{KL}(p_1,p_2) = KL(p_1\|p_2)+KL(p_2\|p_1)
    .}

The KL-divergence approximates the Fisher information distance \cite{Kass&Vos:97},
\Eq{ \label{E:FisherApprox}
\sqrt{D_{KL}(p_1,p_2)}\rightarrow D_F(p_1,p_2)
,}
as $p_1\rightarrow p_2$. If $p_1$ and $p_2$ do not lie closely together on the manifold, this approximation becomes weak, yet a good approximation can still be achieved if the manifold is densely sampled between the two end points. By defining the path between $p_1$ and $p_2$ as a series of connected segments and summing the length of those segments, we approximate the length of the geodesic. Specifically, the Fisher information distance between $p_1$ and $p_2$ can be estimated as:
\Eq{ \label{Eq:FID_Approx_G}
\hat{D}_F(p_1,p_2;\mP)=\min_{m,\mP} {\sum_{i=1}^m{ \sqrt{D_{KL}(p_{(i)},p_{(i+1)})}}}, \quad p_{(i)}\rightarrow p_{(i+1)}\: \forall \: i,}
where $\mP=\left\{p_1,\ldots,p_n\right\}$ is the available collection of PDFs on the manifold. Hence, we are able to use the KL-divergence as a means for calculating similarity between patient data sets. For our purposes, we choose to estimate patient PDFs through kernel density estimation (see Appendix \ref{A:KDE}), although other methods are available (e.g.~mixture models).

\subsection{Objective Function}
In FINE, we found an embedding which mapped information distances between PDFs as Euclidean distances in a low-dimensional space. This allowed us to embed an entire PDF, and therefore all of the cells which were realizations of that PDF, into a single low-dimensional vector. This provided for the direct comparison of patients in the same normalized space. In our current task, we are not interested in embedding a group of patients into the same space, but rather projecting each patient individually in its own space. However, it is important that we maintain differences between patients, as we have found that is a great way to differentiate disease classes.

We define our \emph{Information Preserving Component Analysis (IPCA)} projection as one that preserves the Fisher information distance between data sets. Specifically, let $\mX=\{\vX_1,\vX_2,\ldots,\vX_N\}$ where $\vX_i\in\Real^{d\times n_i}$ is the $n_i$-element data set corresponding to the flow cytometer output of the $i^{th}$ patient, measured with $d$ different markers. We wish to find a single projection matrix $A$ such that
\[
D_{KL}(A\vX_i,A\vX_j)=D_{KL}(\vX_i,\vX_j),\: \forall \: i,j
.\]
Formatting as an optimization problem, we would like to solve:
\Eq{\label{E:proj_obj}
A=\arg\min_{A:AA^T=I} \| D(\mX)-D(\mX,A)\|_F^2
,}
where $I$ is the identity matrix, $D(\mX)$ is a dissimilarity matrix such that $D_{ij}(\mX)=D_{KL}(\vX_i,\vX_j)$, and $D(\mX,A)$ is a similar matrix where the elements are perturbed by $A$, i.e.~$D_{ij}(\mX,A)=D_{KL}(A\vX_i,A\vX_j)$.

Since pathologists view projections in order diagnose based on similar marker expression patterns, maintaining similarities within disease class (and differences between class) is of the utmost importance. These measures are expressed quantitatively through information. By finding the projection solving the objective function (\ref{E:proj_obj}), we ensure that the amount of information between patients which is lost due to the projection is minimized.
\subsection{Gradient Descent}
Gradient descent (or the method of \emph{steepest} descent) allows for the solution of convex optimization problems by traversing a surface or curve in the direction of greatest change, iterating until the minimum is reached. Specifically, let $J(x)$ be a real-valued objective function which is differentiable about some point $x_i$. The direction in which $J(x)$ decreases the fastest, from the point $x_i$, is that of the negative gradient of $J$ at $x_i$, $-\frac{\partial}{\partial x}J(x_i)$. By calculating the location of the next iteration point as
\[
x_{i+1}=x_i-\mu\frac{\partial}{\partial x}J(x_i)
,\]
where $\mu$ is a small number regulating the step size, we ensure that $J(x_i)\geq J(x_{i+1})$. Continued iterations will result in $J(x)$ converging to a local minimum. Gradient descent does not guarantee that the process will converge to an absolute minimum, so typically it is important to initialize $x_0$ near the estimated minimum.

Using gradient descent, we are able to solve (\ref{E:proj_obj}). Specifically, let $J=\| D(\mX)-D(\mX,A)\|_F^2$ be our objective function, measuring the error between our projected subspace and our full-dimensional space. The direction of the gradient is solved by taking the partial derivative of $J$ w.r.t.~a projection matrix $A$,
\[
\frac{\partial}{\partial A}J=\sum_i\sum_j \frac{\partial}{\partial A} \left[D_{ij}(\mX,A)^2-2D_{ij}(\mX)D_{ij}(\mX,A)\right]
.\]
Given the direction of the gradient, the projection matrix can be updated as
\Eq{ \label{E:proj_update}
A=A-\mu\frac{\partial}{\partial A}\tilde{J}(A)
,}
where \[\frac{\partial}{\partial A}\tilde{J}=\frac{\partial}{\partial A}J-\frac{1}{2}\left(\left(\frac{\partial}{\partial A}J\right) A^T+A\left(\frac{\partial}{\partial A}J\right)^T\right)A\] is the direction of the gradient, constrained to force $A$ to remain orthonormal (the derivation of this constraint can be found in Appendix \ref{A:Orth}). This process is iterated until the error $J$ converges.
\subsection{Algorithm}
\begin{algorithm}[t]
\caption{Information Preserving Component Analysis}
\label{a:process}
    \begin{algorithmic}[1]
        \REQUIRE Collection of data sets $\mX=\{\vX_1,\vX_2,\ldots,\vX_N\}$, $\vX_i\in\Real^{d\times n_i}$; the desired projection dimension $m$; search step size $\mu$
        \STATE Calculate $D(\mX)$, the Kullback-Leibler dissimilarity matrix
        \STATE Initialize $A_1\in\Real^{m\times d}$ as a random orthonormal projection matrix
        \STATE Calculate $D(\mX,A_i)$, the Kullback-Leibler dissimilarity matrix in the projected space
        \FOR{$i=1$ to $\infty$}
            \STATE Calculate $\frac{\partial}{\partial A_i}\tilde{J}$, the direction of the gradient, constrained to $AA^T=I$
            \STATE $A_{i+1}=A_i-\mu \frac{\partial}{\partial A_i}\tilde{J}$
            \STATE Calculate $D(\mX,A_{i+1})$
            \STATE $J=\| D(\mX)-D(\mX,A_{i+1})\|_F^2$
            \STATE Repeat until convergence of $J$
        \ENDFOR
        \ENSURE Projection matrix $A\in\Real^{m\times d}$, which preserves the information distances between sets in $\mX$.
    \end{algorithmic}
\end{algorithm}

The full method for IPCA is described in Algorithm \ref{a:process}. We note that $A$ is initialized as a random orthonormal projection matrix due to the desire to not bias the estimation. While this may result in finding a local minimum rather than an absolute minimum, experimental results have shown that the flow cytometry problem is sufficiently convex, at least for our available data, yielding significantly similar convergence values. At this point we stress that we utilize gradient descent due to its ease of implementation. There are more efficient methods of optimization, but that is out of the scope of the current contribution and is an area for future work.

\section{Simulations}
\label{s:results}

\subsection{Synthetic Data}

\begin{figure}[t]
  \center
    \includegraphics[scale=.55]{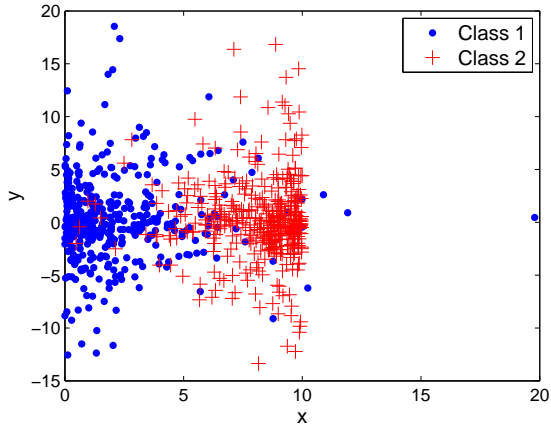}\\
  \caption{An illustration of a sample data set from each class for our synthetic data test. The classes are distributed as `mirror images' of each other, about the line $x=5$.}
  \label{f:synth_sample}
\end{figure}

As a proof of concept, we now illustrate IPCA on a synthetic data set of known structure. An illustration of the data is shown in Fig.~\ref{f:synth_sample}, which is defined as follows: Let $\mX=\left\{\vX_1,\ldots,\vX_{N_1},\vX_{N_1+1},\ldots,\vX_{N_1+N_2}\right\}$ be a collection of sets in which $\vX_j\in\Real^{2\times400}$ is created by joining two Chi-squared distributions (one flipped about the $x$-axis). For $j=1,\ldots,N_1$, let us define $\mX_1$ in that fashion while we define $\mX_2$ for $j=N_1+1,\ldots,N_1+N_2$ in a similar manner, with the data flipped about the $y$-axis and offset by $+10$ units. Essentially, $\mX_1$ and $\mX_2$ contain `mirror image' data sets ('mirrored' about the line $x=5$) with $400$ samples each. We wish to find the projection down to a single dimension which optimally preserves the Fisher information between data sets. For this simulation, let $N_1=N_2=5$.

\begin{figure}[t]
  \center
    \includegraphics[scale=.55]{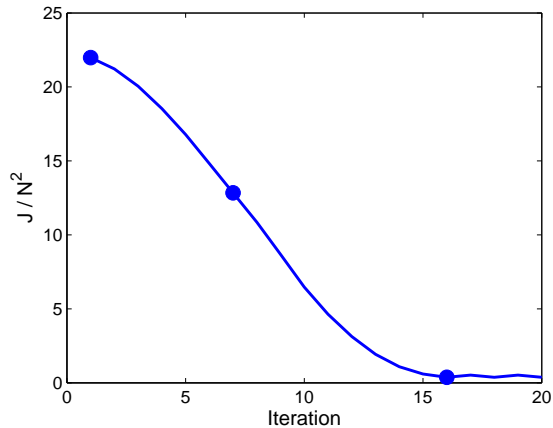}\\
  \caption{The objective function is minimized as we use IPCA to search for the best projection. The circled points correspond to the projections used in Figure \ref{f:synth_evol}.}
  \label{f:synth_error}
\end{figure}

Starting with $A_1\in\Real^{1\times2}$ as a random orthonormal projection matrix, we use IPCA to obtain a projection matrix. Figure \ref{f:synth_error} shows the value of the objective function (normalized to a \emph{per pair} value) as a function of gradient descent iterations. Once the objective function converges, we obtain the projection matrix $A\in\Real^{1\times2}$. This matrix is used to project the data from the $2$ original dimensions down to a dimension of $1$, such that $y_j=A\vX_j$.
\begin{figure*}[t]
  \centerline{
  \subfigure[$A_1$: Random]{
    \includegraphics[scale=.36]{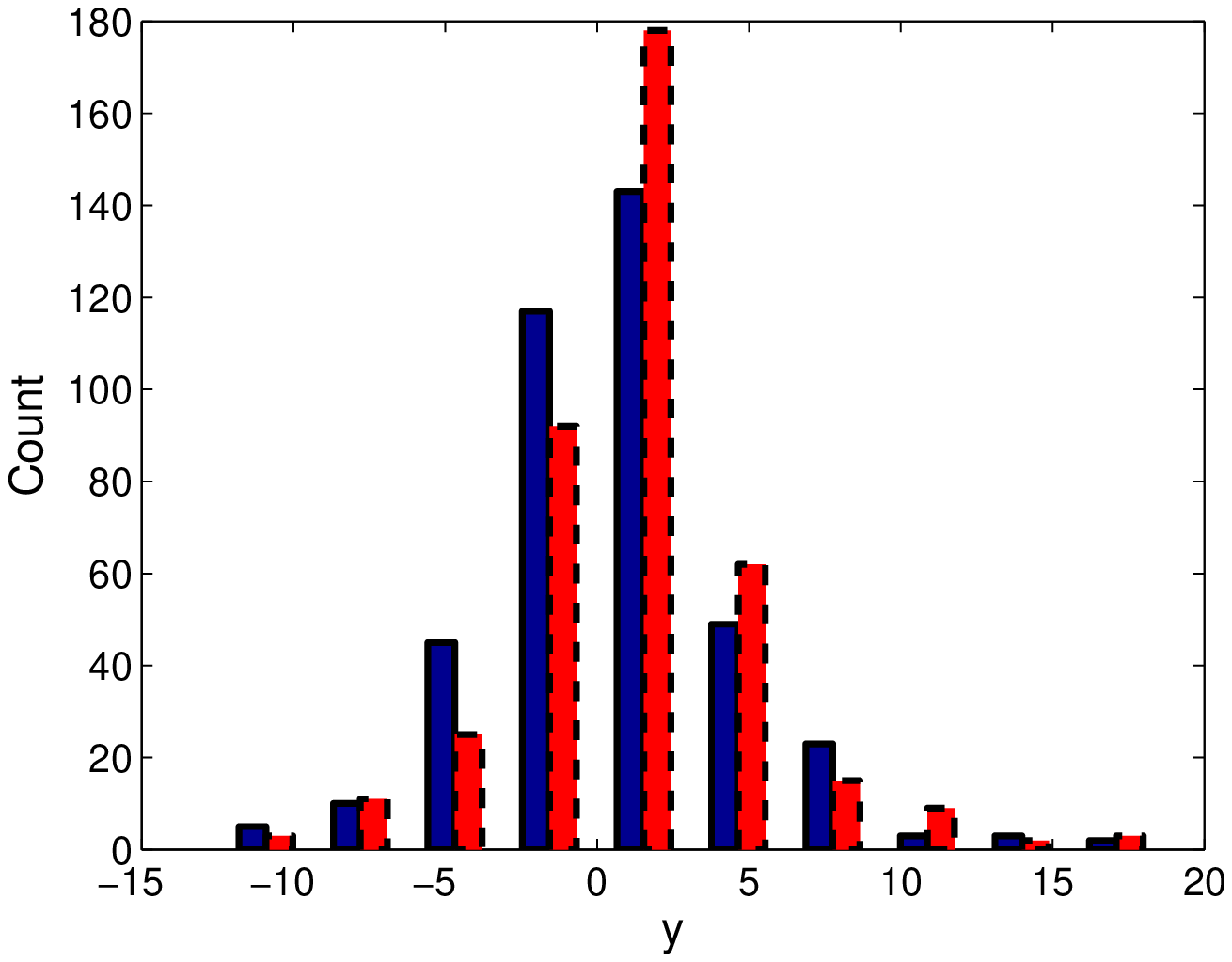}\\}
  \subfigure[$A_7$: Improved]{
    \includegraphics[scale=.36]{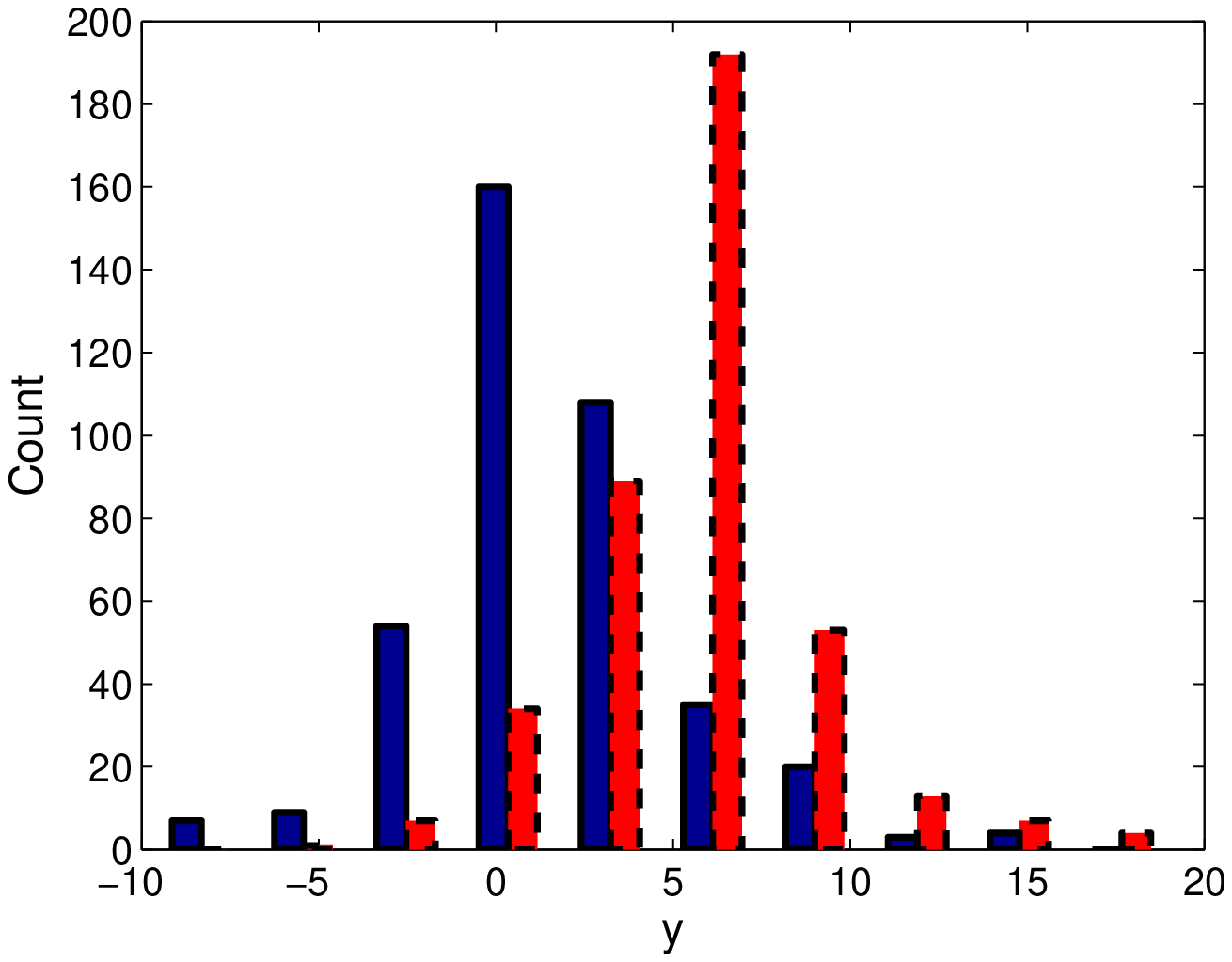}\\}
  \subfigure[$A_{16}$: Best] {\label{f:evol_opt}
    \includegraphics[scale=.36]{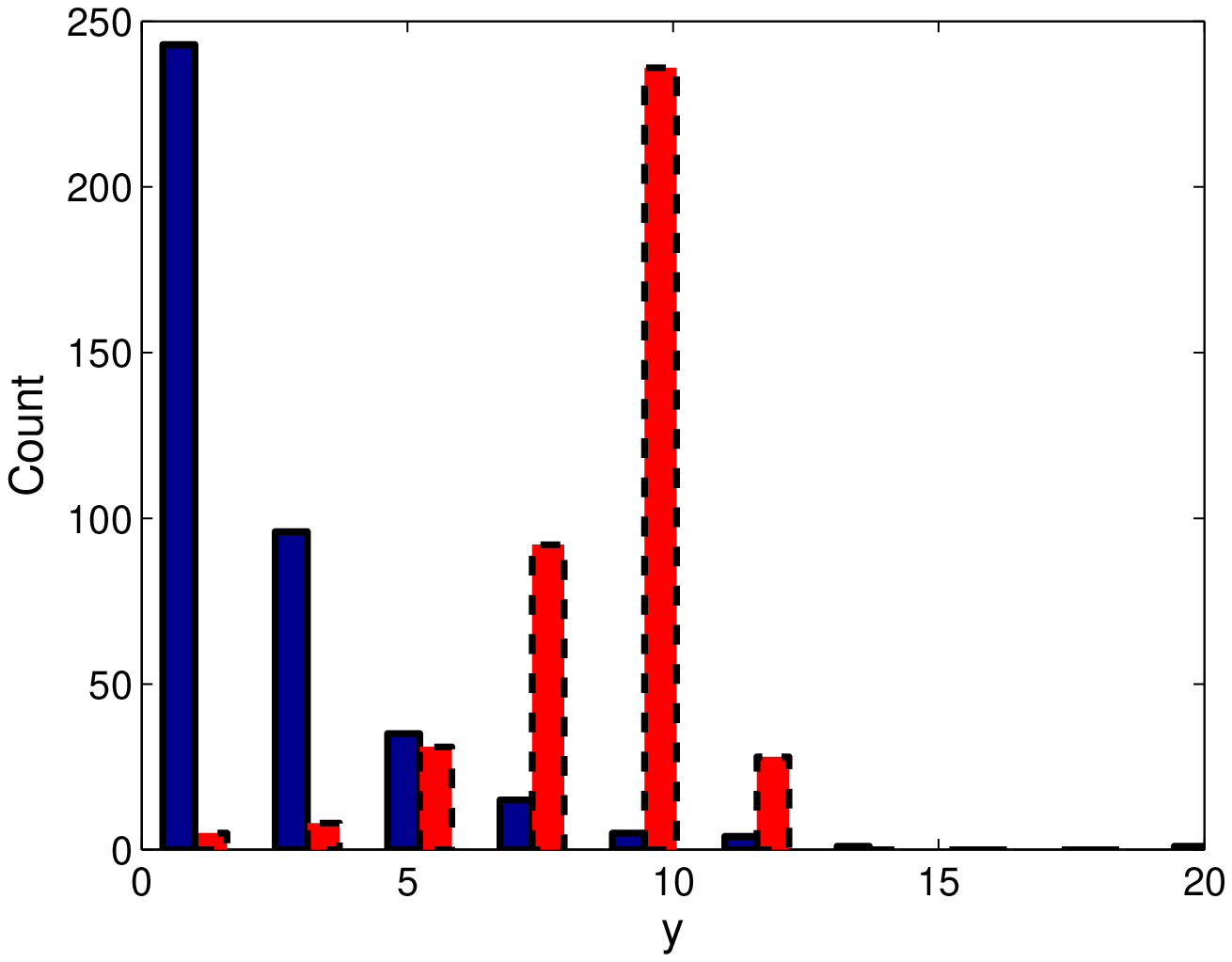}\\}
  }
  \caption{The evolution of the projection matrix, illustrated on one set from each class. As the objective function is minimized, the statistical separation between sets from differing clusters is increased.}
  \label{f:synth_evol}
\end{figure*}


The evolution of the projection matrix is illustrated in Fig.~\ref{f:synth_evol}. One set from each cluster was projected onto the $1$-dimensional space defined by $A_i$ (as highlighted in Fig.~\ref{f:synth_error}). The initial projection matrix $A_1$, which was randomly generated, offers no distinction between the sets from differing clusters. As the algorithm searches to minimize the objective function, the projection matrix begins to recognize structure within the data, and the sets begin to separate. This process continues until the best projection matrix (in this case $A_{16}$) is found and the sets are well distinguished. We stress that the distinguishing characteristic is not the Euclidean location of the samples within each data set (as we see they contain some overlap), but the statistics of each set. 


\begin{table}[t!]
\center{
\begin{tabular}{|l|c|c|}
\hline
  & \multicolumn{2}{c|} {Objective Function} \\ \hline
  Method & Mean & Standard Deviation \\ \hline

  IPCA & \textbf{0.4006} & \textbf{0.1687} \\
  ICA & 0.4539 & 0.2204 \\
  PCA & 22.7837 & 2.3850 \\
  \hline
\end{tabular}}
\caption{Comparing the performance of IPCA to that of PCA and ICA over a 10-fold cross validation. The IPCA projection method outperforms both other methods.}
\label{t:us_v_them}
\end{table}

We compare our methods to those of well known linear and unsupervised dimensionality reduction algorithms, namely Principal Components Analysis (PCA) and Independent Component Analysis (ICA). To obtain the PCA and ICA projection matrices, we combined all data into a single set, then performed the corresponding algorithm on that set, defining the projection matrix as the first $m$ principal/independent components ($m=1$ in this case). The results over a $10$-fold cross validation are demonstrated in Table \ref{t:us_v_them}, where IPCA shows superior performance to comparative methods. PCA performs particularly poorly as it projects data onto the directions with the highest variance. It does not recognize the inherent structure in data which contains interesting properties in directions of low variance, such as this example. For ICA we selected the component with the highest norm \cite{Fodor:02}, as the ICs are not ordered \cite{Hyvarinen&Oja:NN00}. This method performs admirably well, correctly identifying the direction of interest, although still falling short of the performance of IPCA.

\subsubsection{Variable Selection}
One immediately noticeable benefit of IPCA is that we may use the loading vectors of $A$ towards the problem of variable selection. IPCA finds the linear combination of channels which best preserves the information between data sets. Given the definition of the Kullback-Leibler divergence (\ref{Equation:KL}), the dimensions which contribute most to the information are those in which data sets differ most in probability distribution. As such, the loading vectors in $A$ will be weighted towards the most discriminating variables.

Continuing with our previous example, the IPCA projection matrix was always of the order $A=[1-\epsilon_1, \epsilon_2]$, in which $\epsilon_{1,2}\approx0$. This result is obvious, as all of the data sets are identically distributed within the second dimension and the only differentiating variable is the first dimension. While this result is trivial for our synthetic example, the ability to use the loading vectors as a means of variable selection shall prove vital when applied to real-world data.

\subsection{Flow Cytometry Analysis}
We now present simulation results for using IPCA to find a projection matrix for flow cytometric data analysis. We demonstrate three distinct studies involving differing disease classes to show that our methods are not just beneficial to a single example. We offer a proof of concept that shall allow pathologists to utilize our methods on many different studies and for exploratory data analysis. In all cases, patient data was obtained and diagnosed by the Department of Pathology at the University of Michigan.

\subsubsection{Lymphoid Leukemia Study}
For our first study, we will compare patients with two distinct but immunophenotypically similar forms of lymphoid leukemia -- mantle cell lymphoma (MCL) and chronic lymphocytic leukemia (CLL). These diseases display similar characteristics with respect to many expressed surface antigens, but are generally distinct in their patterns of expression of two common B lymphocyte antigens: CD23 and FMC7. Typically, CLL is positive for expression of CD23 and negative for expression of FMC7, while MCL is positive for expression of FMC7 and negative for expression of CD23.  These distinctions should lead to a difference in densities between patients in each disease class.

\begin{table}[t]
\center{
\begin{tabular}{|l|l|}
  \hline
  Dimension & Marker \\
  \hline
  1 & Forward Light Scatter \\
  2 & Side Light Scatter \\
  3 & FMC7 \\
  4 & CD23 \\
  5 & CD45 \\
  6 & Empty \\
  \hline
\end{tabular}}
\caption{Data dimensions and corresponding markers for analysis of CLL and MCL.}
\label{t:cll_mcl_markers}
\end{table}

The data set $\mX=\left\{\vX_1,\ldots,\vX_{43}\right\}$ consists of 43 patients, 23 of which have been diagnosed with CLL and 20 diagnosed with MCL. Each $\vX_i$ is a $6$ dimensional matrix, with each dimension corresponding to a different marker (see Table \ref{t:cll_mcl_markers}), and each element representing a unique blood cell, totaling $n_i\sim5000$ total cells per patient. We calculate $D(\mX)$, the matrix of Kullback-Leibler similarities, and desire to find the projection matrix $A$ that will preserve those similarities when all data sets are projected to dimension $d=2$.

\begin{figure}[t]
  \centerline{
  \includegraphics[scale=.55]{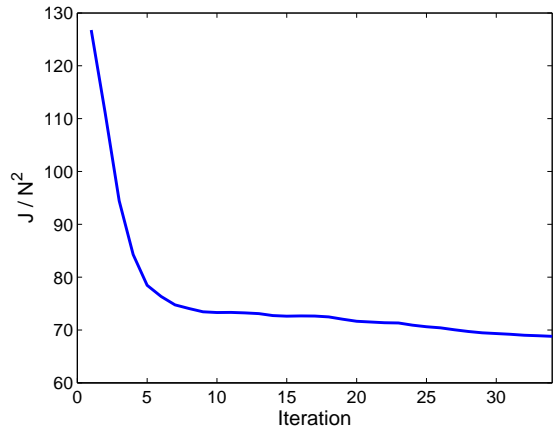}}
  \caption{CLL and MCL Study: Evaluating the objective as a function of time. As the iterations increase, the objective function eventually converges.}
  \label{f:cll_mcl_J}
\end{figure}

Using the methods described in this paper, we found the IPCA projection as
\Eq{\label{e:A_cll}
A=\left(
    \begin{array}{cccccc}
      0.0640 & 0.0364 & 0.9055 & 0.2075 & 0.3547 & -0.0842 \\
      -0.0188 & -0.1969 & -0.1453 & -0.9557 & 0.1646 & -0.0111 \\
    \end{array}
  \right)
.}
This projection was calculated by minimizing the objective function with respect to $A$, as illustrated in Fig.~\ref{f:cll_mcl_J} in which the squared error (per element pair) is plotted as a function of time. As the iteration $i$ increases, $J$ converges and $A_i$ is determined to be the IPCA projection matrix. We note that while dimension 6 corresponds to no marker (it is a channel of just noise), we do not remove the channel from the data sets, as the projection determines this automatically (i.e.~loading values approach $0$). Additionally, due to computational complexity issues, each data set was randomly subsampled such that $n_i=500$. While we would not suggest this decimation in practice, we have found it to have a minimal effect during experimentation.

\begin{figure}[t]
  \centerline{
  \subfigure[Most Similar] {
  \begin{tabular}{c}
    \includegraphics[scale=.35]{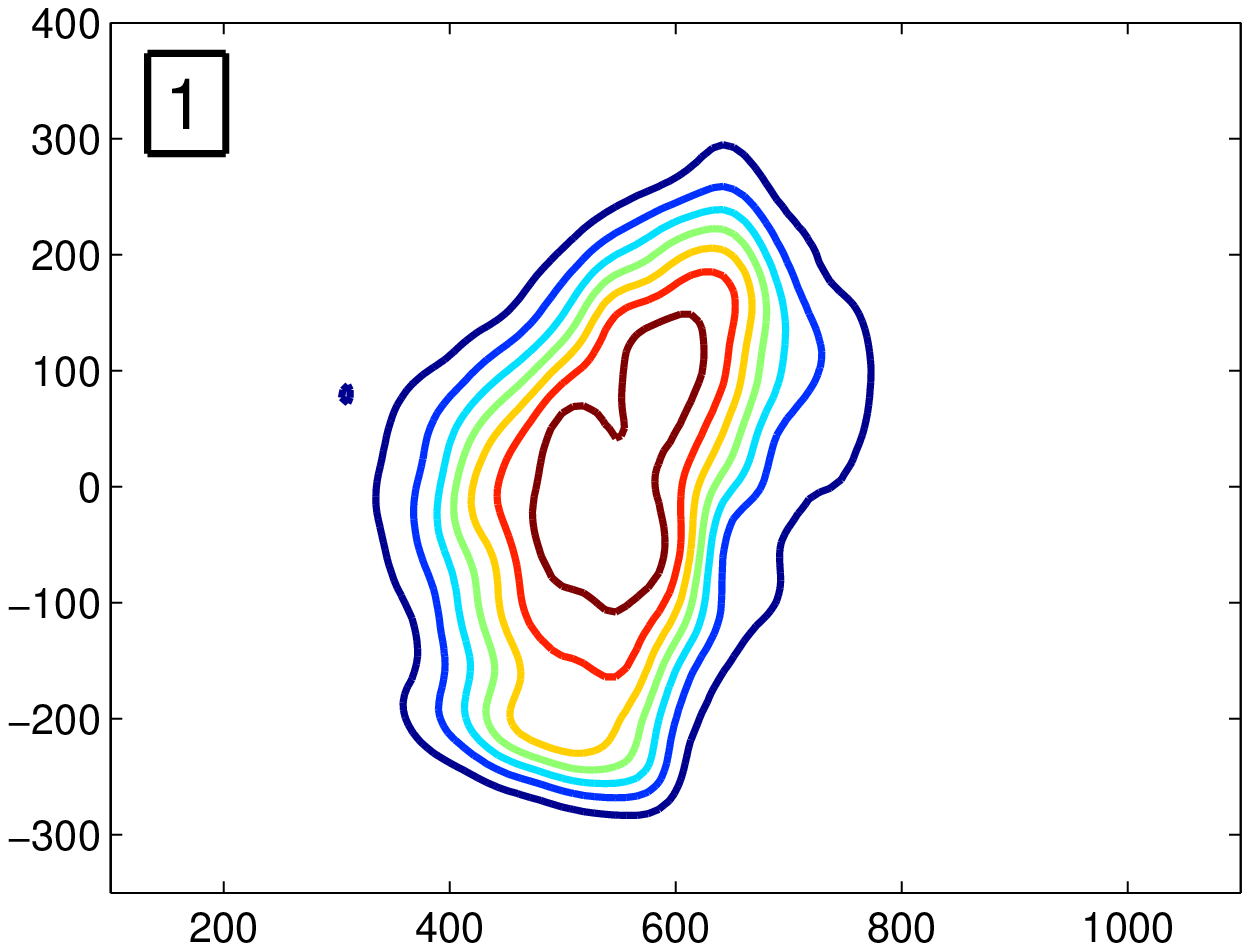}\\
    \includegraphics[scale=.35]{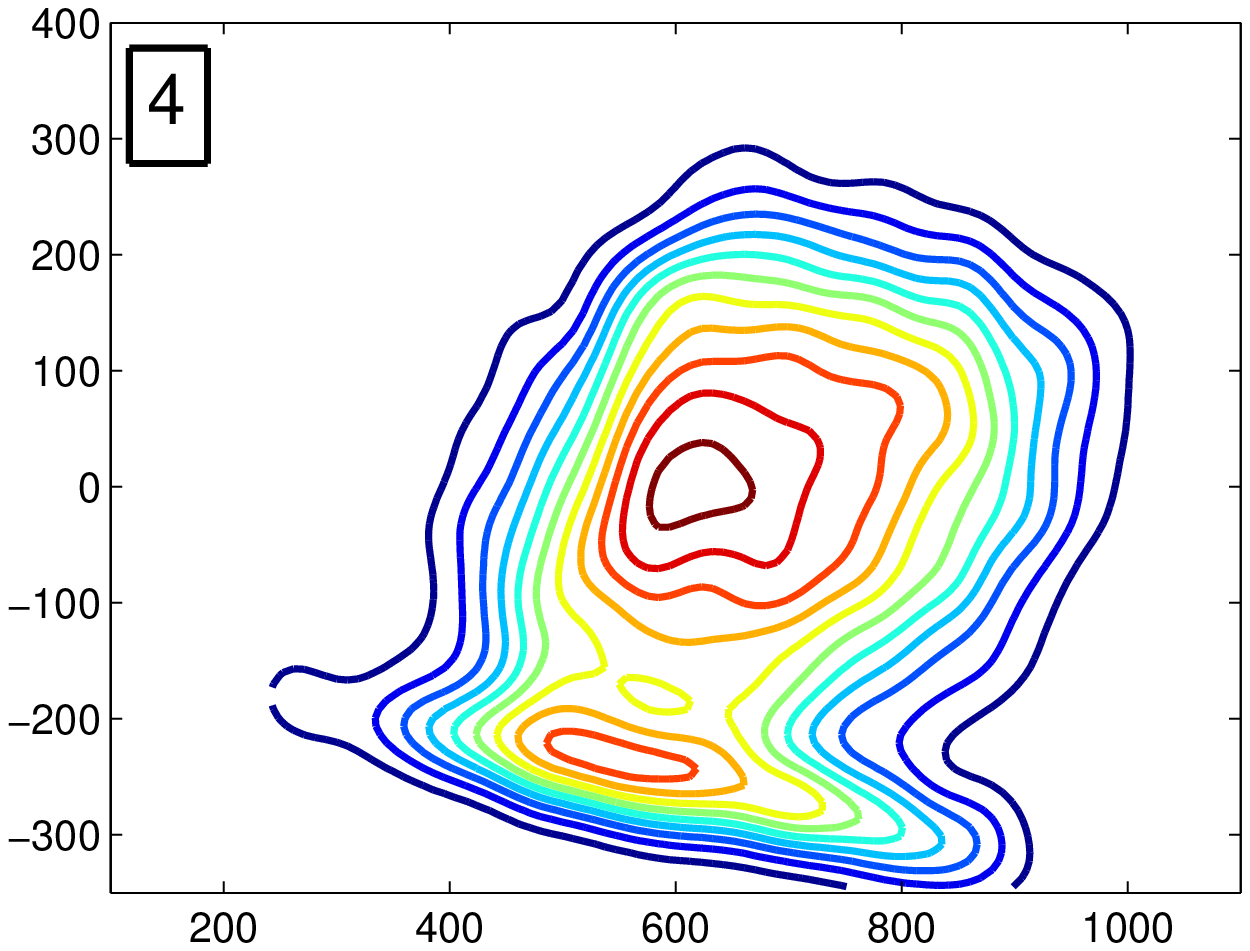}\\
  \end{tabular}
  }
  \subfigure[Centroids] {
  \begin{tabular}{c}
    \includegraphics[scale=.35]{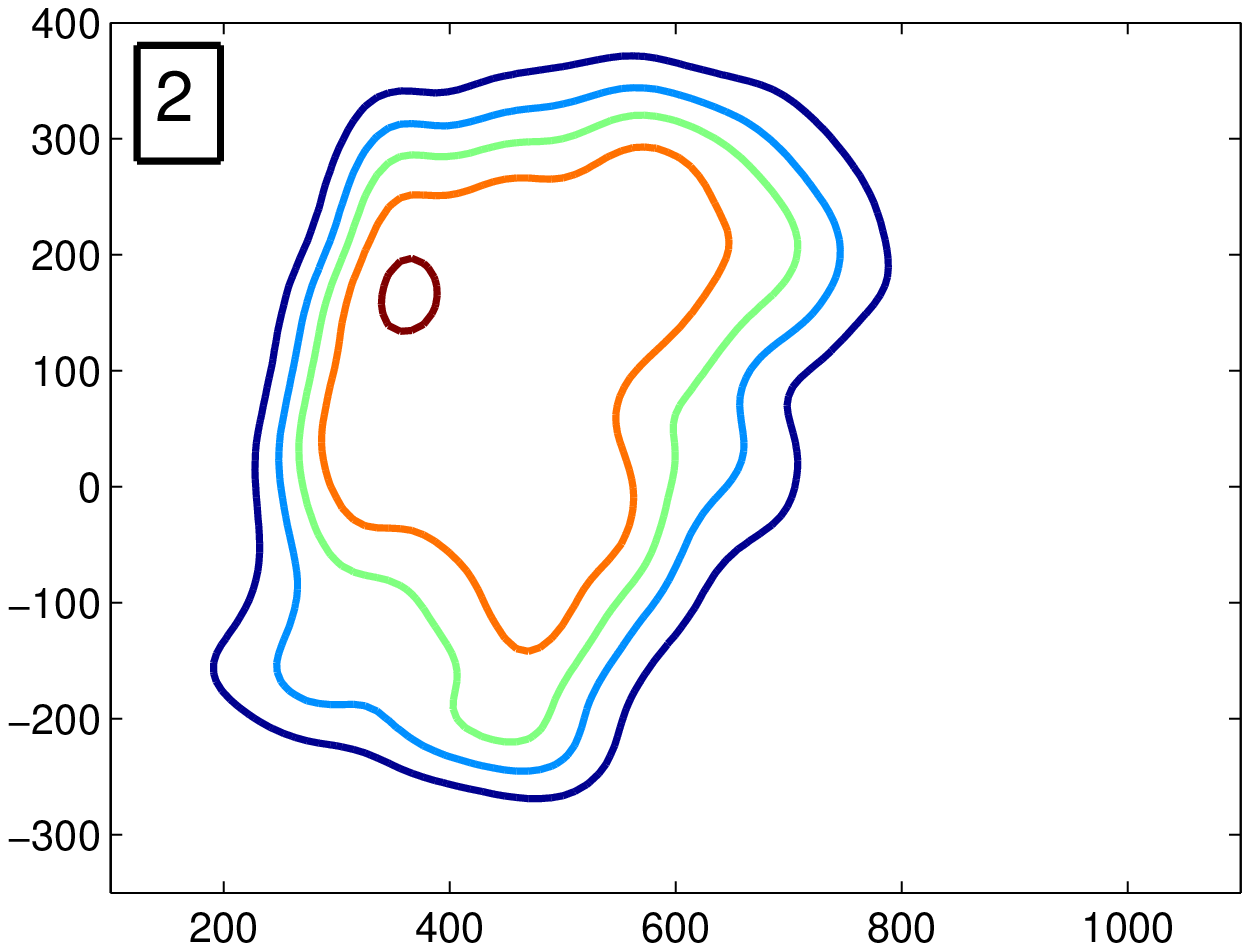}\\
    \includegraphics[scale=.35]{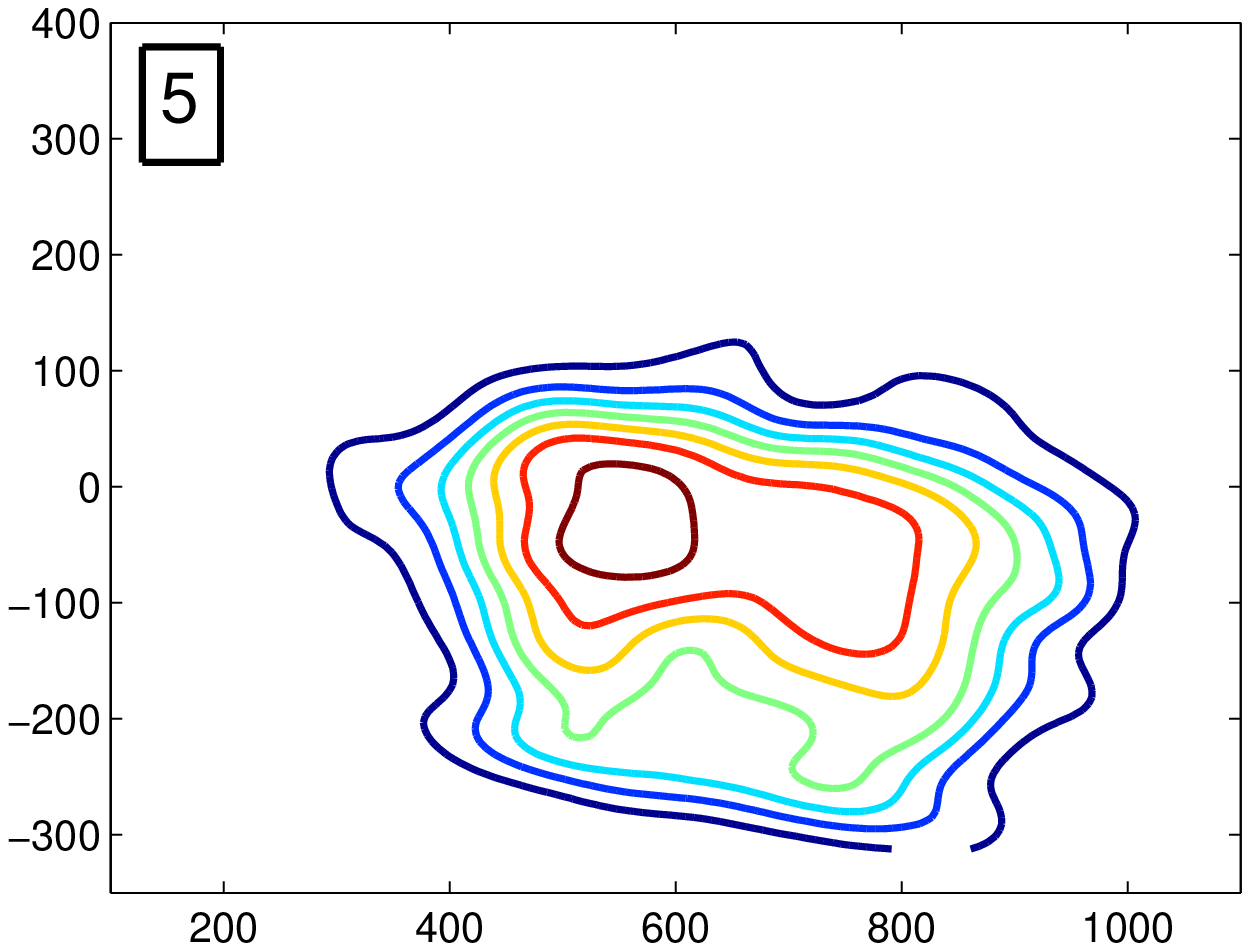}\\
  \end{tabular}
  }
  \subfigure[Least Similar] {
  \begin{tabular}{c}
    \includegraphics[scale=.35]{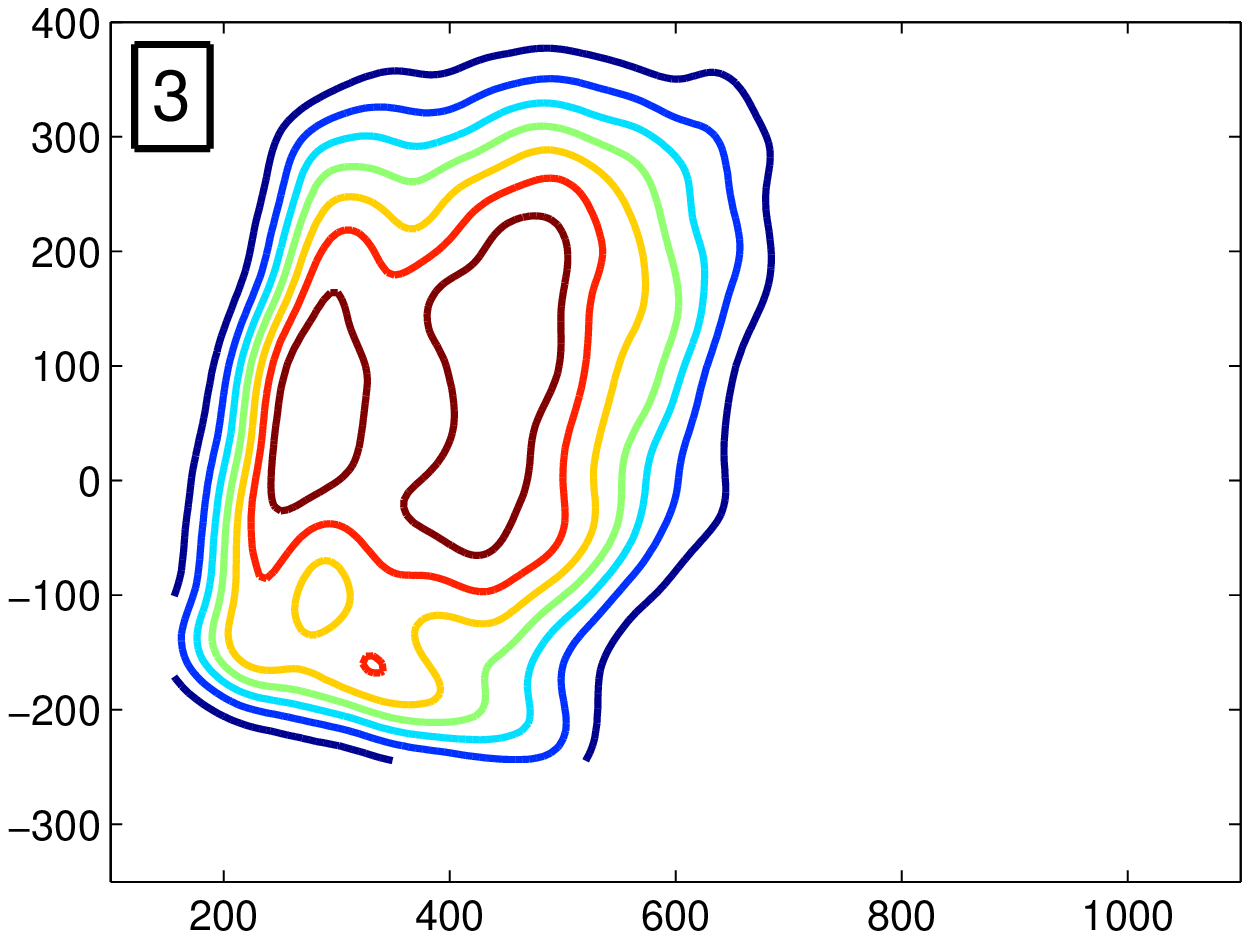}\\
    \includegraphics[scale=.35]{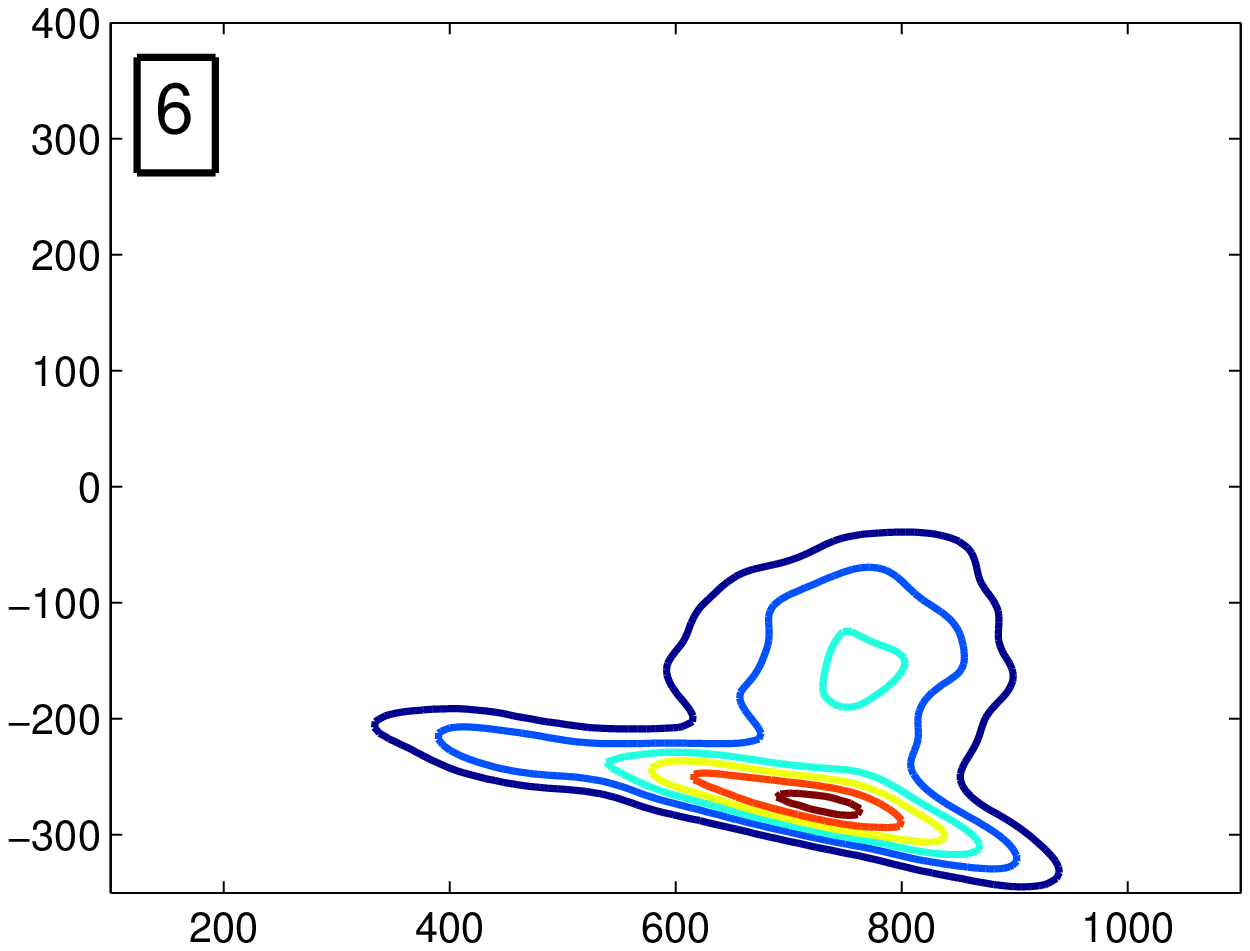}\\
  \end{tabular}
  }
  }
  \caption{CLL and MCL Study: Contour plots (i.e.~PDFs) of the IPCA projected data. The top row corresponds to the PDFs the CLL patients, while the bottom row represents PDFs of MCL patients. The selected patients are those most similar between disease classes, the centroids of disease classes, and those least similar between disease classes, as highlighted in Fig.~\ref{f:fine_opt}.}
  \label{f:dens_proj}
\end{figure}

Given the IPCA projection, we illustrate the 2-dimensional PDFs of several different patients in the projected space in Fig.~\ref{f:dens_proj}. We selected patients based on the KL-divergence values between patients of different disease class. Specifically, we selected the CLL and MCL patients with a small divergence (i.e.~most similar PDFs), patients with a large divergence (i.e.~least similar PDFs), and patients which represented the centroid of each disease class. These low-dimensional PDFs, which are what would be utilized by a diagnostician, are visibly different between disease classes. While the most similar CLL and MCL patients do share much similarity in their IPCA PDFs, there is still a significant enough difference to distinguish them, especially given the similarities to other patient PDFs.

\begin{figure}[t]
  \centerline{
  \subfigure[Full Dimension] {\label{f:fine_full}
    \includegraphics[scale=.55]{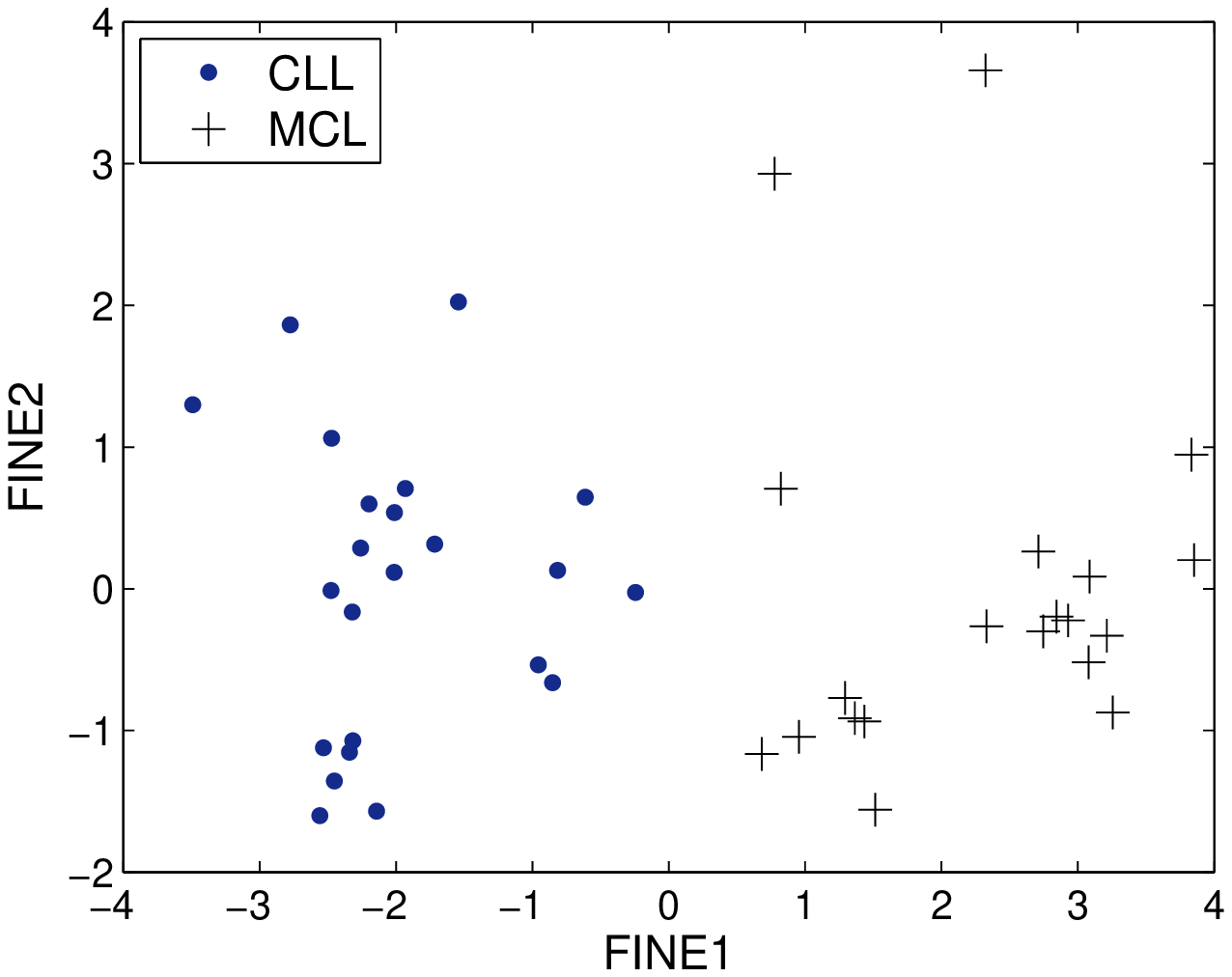}\\
  }
  \subfigure[IPCA Projection] {\label{f:fine_opt}
    \includegraphics[scale=.55]{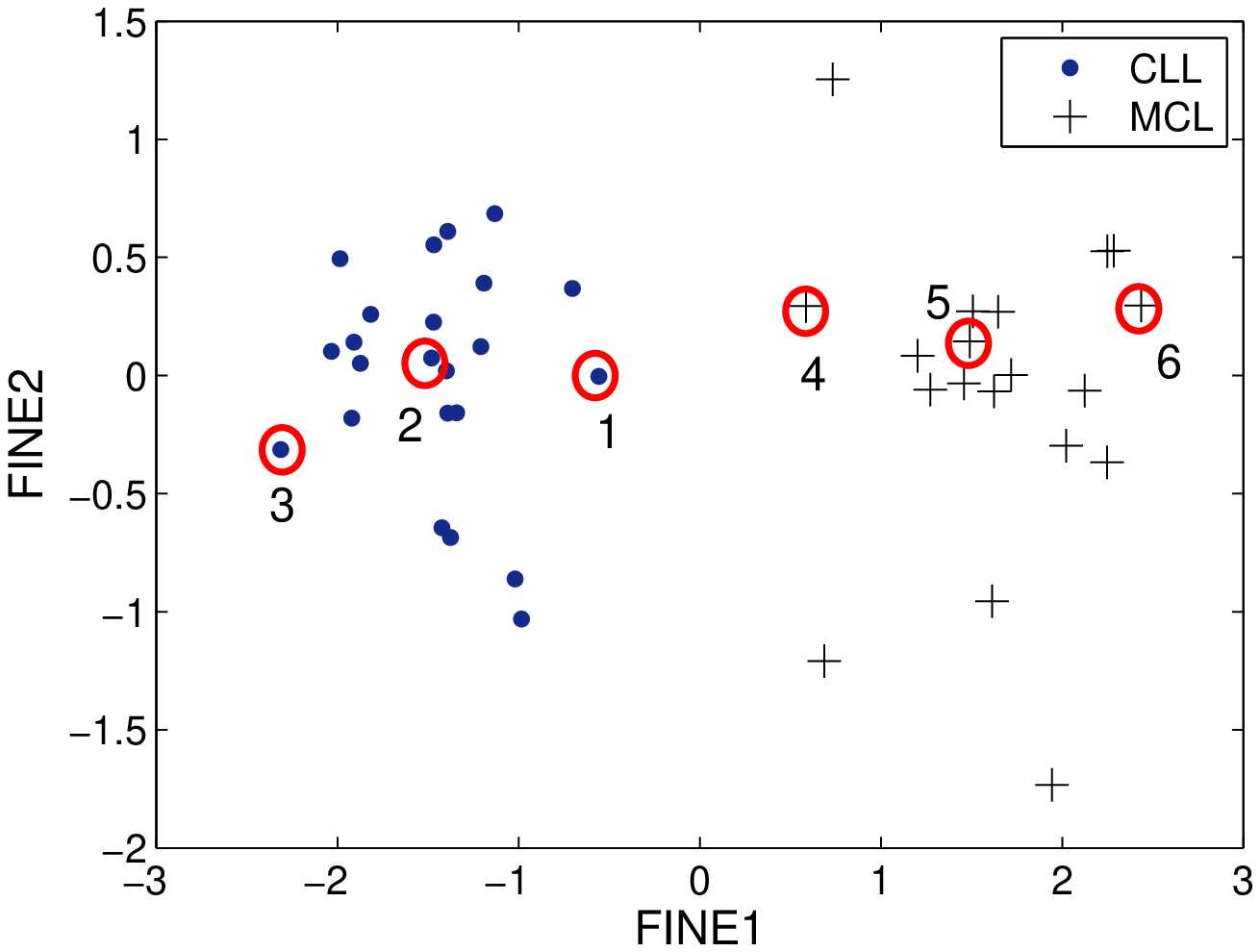}\\}
  }
  \caption{CLL and MCL Study: Comparison of embeddings, obtained with FINE, using the full dimensional data and the data projected with IPCA. IPCA preserves the separation between disease classes. The circled points correspond to the density plots in Fig.~\ref{f:dens_proj}, numbered respectively.}
  \label{f:fine_compare}
\end{figure}

We now illustrate the embedding obtained with FINE of the projected data (see Appendix \ref{AP:FINE}). The embedding results are shown in Fig.~\ref{f:fine_opt}, in which the separation between classes is preserved when using the projected data as compared to using the full-dimensional data in Fig.~\ref{f:fine_full}. Each point represents an entire patient data set, and those which are circled correspond to the PDFs shown in Fig.~\ref{f:dens_proj}. By finding the projection which minimizes the difference in KL-divergence between the full and projected data, we maintain the relationships between different sets, allowing for a consistent analysis. 

Using the projection matrix (\ref{e:A_cll}) for variable selection, the loading vectors are highly concentrated towards the 3$\rm^{rd}$ and 4$\rm^{th}$ dimensions, which correspond to fluorescent markers FMC7 and CD23. 
We acknowledge that this marker combination is well known and currently utilized in the clinical pathology community for differentiating CLL and MCL\footnote{CD45 and light scatter characteristics are often used as gating parameters for selection of lymphocytes among other cell types prior to analysis, but CD23 and FMC7 are the main analytical biomarkers in this 3-color assay.}. We stress, however, that what had previously been determined through years of clinical experience was able to be independently validated quickly using IPCA. This is important as it could enable pathologists to experiment with new combinations of fluorescent markers and see which may have strong effects on the discernment of similar leukemias and lymphomas.

\subsubsection{Chronic Lymphocytic Leukemia Study}
\label{ss:cll}
Continuing our study of patients with chronic lymphocytic leukemia (CLL), we wish to determine subclasses within the CLL disease class. Specifically, we now use IPCA to find a low-dimensional space which preserves the differentiation between patients with good and poor prognoses (i.e.~favorable and unfavorable immunophenotypes). Literature \cite{Damle&EtAl:Blood99} has shown that patients whose leukemic cells are strong expressors of CD38 have significantly worse survival outcome. Genotypic studies have shown that the absence of somatic mutation within immunoglobulin genes of CLL cells (a so-called ``pre-follicular'' genotype) is a potent predictor of worse outcome. High levels of CD38 expression are an effective surrogate marker for the absence of somatic immunoglobulin gene mutation, and also have been shown to be an independent predictor of outcome in some studies. Since patients can generally be stratified by CD38 expression levels, and CD38 has been shown to emerge as a defining variable of CLL subsets in hierarchical immunophenotypic clustering \cite{Habib&Finn:CytB06}, we would expect IPCA to localize the CD38 variable as one of importance when analyzing CLL data.

\begin{table}[t]
\center{
\begin{tabular}{|l|l|}
  \hline
  Dimension & Marker \\
  \hline
  1 & Forward Light Scatter \\
  2 & Side Light Scatter \\
  3 & CD5 \\
  4 & CD38 \\
  5 & CD45 \\
  6 & CD19 \\
  \hline
\end{tabular}}
\caption{Data dimensions and corresponding markers for analysis of CLL.}
\label{t:cll_markers}
\end{table}

\begin{figure}[t]
  \centerline{
  \includegraphics[scale=.55]{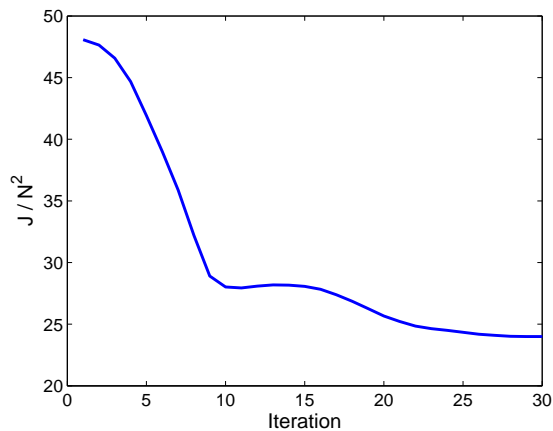}}
  \caption{CLL Prognosis Study: The value of the objective function vs.~time.}
  \label{f:cll_error}
\end{figure}

Using the same patients (those diagnosed with CLL) as in the above simulation, we define $\mX=\left\{\vX_1,\ldots,\vX_{23}\right\}$, where each $\vX_i$ was analyzed with by the series of markers in Table \ref{t:cll_markers}. Minimizing the objective function (see Fig.~\ref{f:cll_error}), we calculate the IPCA projection matrix as
\[
A=\left(
    \begin{array}{cccccc}
      -0.2328 & -0.1160 & -0.3755 & 0.1789 & 0.4615 & 0.7401 \\
      0.1133 & -0.1291 & -0.2712 & 0.8100 & -0.4948 & -0.0064 \\
    \end{array}
  \right)
.\]
This matrix has very high loadings in variables $4$ and $6$, which correspond to markers CD38 and CD19 respectively, corresponding to the isolation of B cells by CD19 expression (a B lymphocyte restricted antigen always expressed on CLL cells) and assessment of CD38 on these B cells. As expected, we identify CD38 as a marker of importance in differentiating patient groups. We also identify the possibility that CD19 expression as an area which may help prognostic ability. This is an area for further interrogation.

\begin{figure}[t]
  \centerline{
  \subfigure[Full Dimensional] {
    \includegraphics[scale=.55]{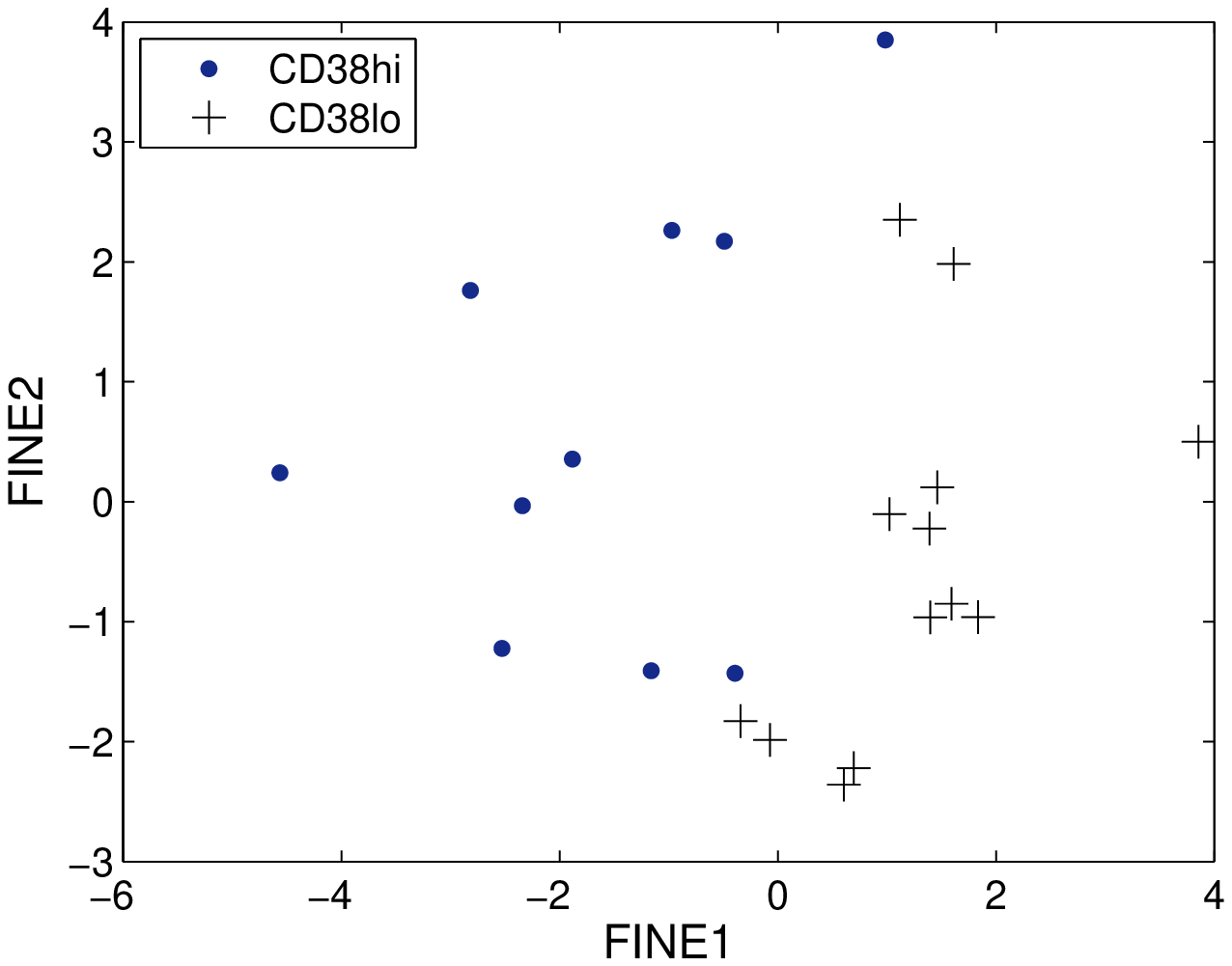}\\
  }
  \subfigure[IPCA Projection] {
    \includegraphics[scale=.55]{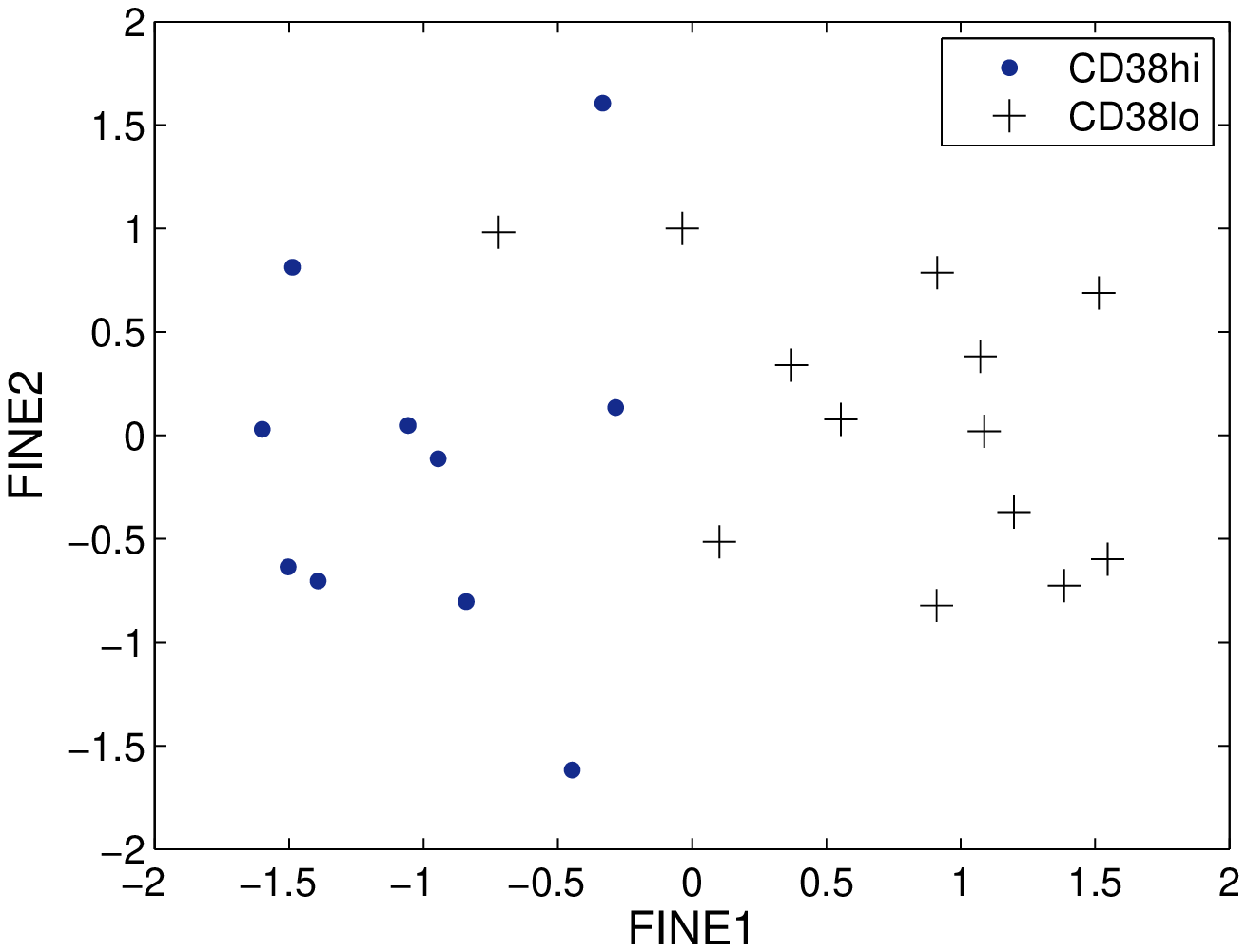}\\}
  }
  \caption{CLL Prognosis Study: Comparison of embeddings, obtained with FINE, using the IPCA projection matrix $A$ and the full dimensional data. The patients with a poor prognosis (CD38hi) are generally well clustered against those with a favorable prognosis (CD38lo) in both embedddings.}
  \label{f:cll_fine}
\end{figure}

Using FINE to embed the data (Fig.~\ref{f:cll_fine}) for comparative visualization, we see that the IPCA projection preserves the grouping of patients with unfavorable immunophenotype (CD38hi) and favorable immunophenotype (CD38lo). CD38hi versus CD38lo for each patient was determined using cutoff values endorsed in the literature \cite{Damle&EtAl:Blood99}. Although complete follow-up data for this retrospective cohort were not available, the findings were indirectly further validated by the fact that, of the patients with follow-up information available, zero of 6 CD38lo patients died, while 4 of 9 CD38hi patients died within a median follow-up interval of 25 months (range 1 to 102 months). As such, we find that IPCA can locate sub-classes and may be useful for possible help towards prognosis.


\subsubsection{Acute Lymphoblastic Leukemia vs. Hematogone Hyperplasia Study}
We now demonstrate a study involving the diseases acute lymphoblastic leukemia (ALL) and a benign condition known as hematogone hyperplasia (HP). ALL is marked by the neoplastic proliferation of abnormal lymphocyte precursors (lymphoblasts). Our study specifically focused upon ALL consisting of B cell precursor lymphobalsts (B-precursor ALL), the most common form of this disease, since the normal counterpart to B-precursor lymphoblasts, termed hematogones, are detectable in the bone marrow of most healthy individuals, and hematogones can proliferate in benign reversible fashion in numerous clinical states \cite{McKenna&EtAl:Blood01}. The distinction between hematogones and leukemic B-precursor lymphoblasts is highly relevant in clinical practice since these cell types exhibit substantial immunophenotypic overlap, many transient conditions associated with hematogone hyperplasia can present with clinical suspicion for leukemia, and patients with ALL can develop HP during recovery from chemotherapy for their leukemia.

For this study, let us define the data set $\mX=\left\{\vX_1,\ldots,\vX_{54}\right\}$, which consists of 54 patients, 31 of which have been diagnosed with ALL and 23 diagnosed with HP. Patient samples were analyzed with a series of markers (see Table \ref{t:all_markers}) designed for the isolation of hematogones and aberrant lymphoblast populations, based on known differential patterns of these markers in these cell types. Specific details of how the data was retrieved can be found in \cite{Finn&Carter:CytB08}.

\begin{table}[t]
\center{
\begin{tabular}{|l|l|}
  \hline
  Dimension & Marker \\
  \hline
  1 & Forward Light Scatter \\
  2 & Side Light Scatter \\
  3 & CD38 \\
  4 & CD19 \\
  5 & CD45 \\
  6 & CD10 \\
  \hline
\end{tabular}}
\caption{Data dimensions and corresponding markers for analysis of ALL and HP.}
\label{t:all_markers}
\end{table}

\begin{figure}[t]
  \centerline{
  \includegraphics[scale=.55]{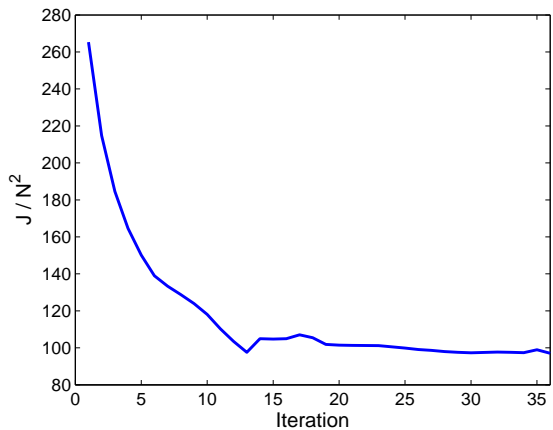}}
  \caption{The value of the objective function (v.s.~time) for the analysis of ALL and HP diagnosis.}
  \label{f:all_v_hp_error}
\end{figure}

\begin{figure}[t]
  \centerline{
  \subfigure[Full Dimensional] {
    \includegraphics[scale=.55]{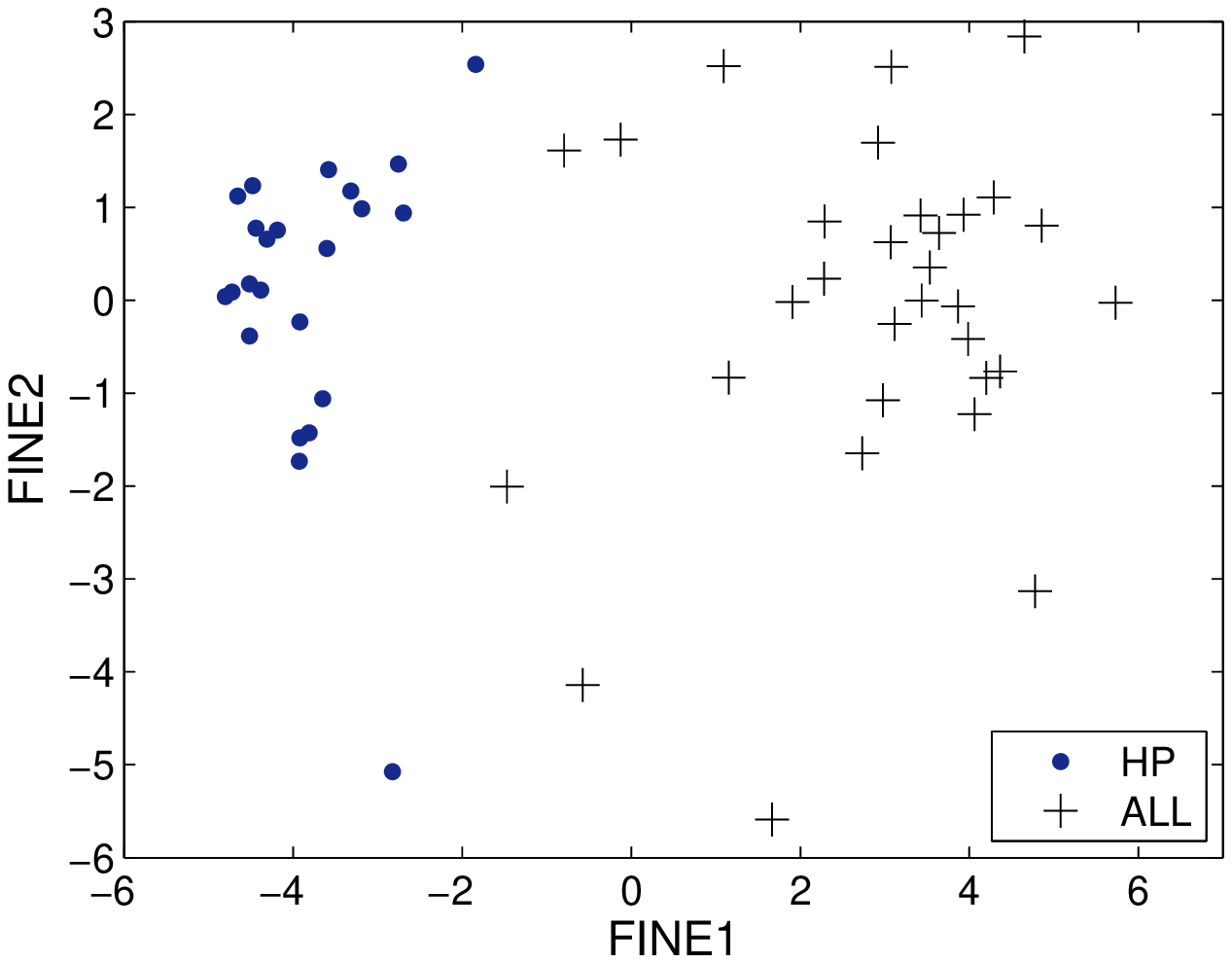}\\
  }
  \subfigure[IPCA Projection] {
    \includegraphics[scale=.55]{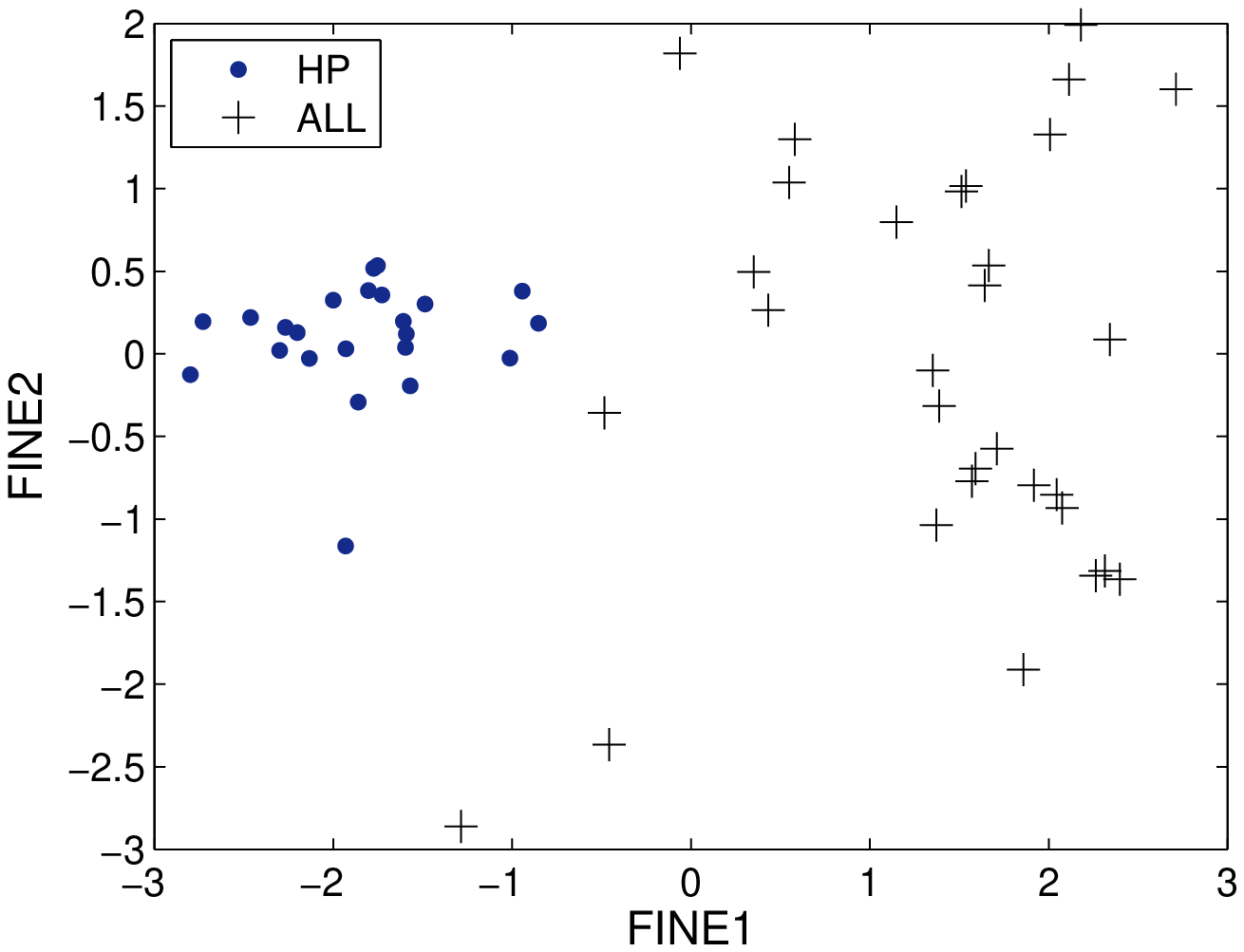}\\}
  }
  \caption{ALL and HP Study: Comparison of embeddings, obtained with FINE, using the full dimensional data and the IPCA projection matrix $A$. The embedding is very similar when using the projected data, which preserves the similarities between patients.}
  \label{f:all_fine}
\end{figure}

By minimizing the objective function (Fig.~\ref{f:all_v_hp_error}), we find the IPCA projection as
\[
A=\left(
    \begin{array}{cccccc}
      0.4041 & 0.2521 & 0.7712 & 0.3795 & 0.4042 & -0.2427 \\
      -0.2665 & -0.2216 & 0.3707 & 0.5541 & -0.3413 & 0.6074 \\
    \end{array}
  \right)
.\]
Using FINE, we compare the embedding of the full-dimensional data to that of the projected data in Fig.~\ref{f:all_fine}. The embeddings are very similar, which illustrates once again that IPCA preserves the similarities between different sets. This allows for a low-dimensional analysis in the projected space with the security of knowing the relationships between patients have been minimally effected.

We also observe that the projection matrix has strong loadings corresponding to markers CD38 and CD10. In clinical practice, it is often noted that hematogones have a very uniform and strong CD38 expression pattern, while lymphoblasts can have quite a range of CD38 expression \cite{McKenna&EtAl:Blood01}. This analysis seems to provide independent validation for that observation. Furthermore, this analysis identifies CD10 as a principal distinguishing marker among the others analyzed in this 4-color assay. This finding is not intuitive, since in day-to-day practice CD10 is not obviously of greater distinguishing value than marker such as CD45 or side angle light scatter. These markers, like CD10, are used for their different expression patterns in lymphoblasts versus hematogones, but that may show considerable overlap in expression intensity between these two cell types. Our identification of CD10 as a marker of importance identifies an area for further clinical investigation.

\section{Conclusions}
\label{s:conclusions}
In this paper we have shown the ability to find an information-based projection for flow cytometric data analysis using Information Preserving Component Analysis (IPCA). By preserving the Fisher information distance between oncological data sets (i.e.~patients), we find a low-dimensional projection that allows for visualization in which the data is discernable between cancerous disease classes. As such, we use machine-learning to provide a projection space that is usable for verification of cancer diagnosis. Additionally, analysis of the loading vectors in the projection matrix allows for a means of variable selection. We have shown independent verification for determining optimal marker combinations in distinguishing immunophenotypically similar cancers, as well as validating variables which help to identify prognostic groups. Verifying these known results through independent methods provides a solid \emph{proof-of-concept} for the ability to utilize IPCA for exploratory research of different marker assays.

In future work we plan to study the effects of preserving only the local distances between data sets. As we have stated, the KL-divergence becomes a weak approximation as the densities separate on the statistical manifold. As such, performance may improve by putting more emphasis on preserving the close distances. However, this may have the adverse effect of diminishing the ability to distinguish between disease classes if they are well separated, as those far distances may not be well preserved.  Additionally, we would like to utilize different methods for optimizing the cost function. While we currently utilize gradient descent for ease of implementation, it is relatively slow and there are more efficient methods to use (ex.~fixed point iteration). The optimization method is not the focus of our work, but faster methods may be required for practical usage. Finally, we would like to apply our methods towards exploratory research and determine other applications of interest.

\appendix
\subsection{FINE Algorithm}
\label{AP:FINE}
We have previously \cite{Carter:TSP08,Carter&Raich:ICASSP08} presented an algorithm for determining a low-dimensional Euclidean embedding of high-dimensional data sets $\mX=\left\{\vX_1,\ldots,\vX_N\right\}$. Coined \emph{Fisher Information Non-parametric Embedding} (FINE), we determine a mapping:
\[
\Psi:\vX_i\rightarrow y_i,\: y_i\in\Real^d
\]
where $\vX_i\in\Real^{m\times n_i}$ is a data set generated by some PDF which exists on an underlying statistical manifold. By approximating the Fisher information distance with a geodesic along the manifold, we can reconstruct a representation of the manifold in Euclidean space, $\vY=\left\{y_1,\ldots,y_N\right\}$. Details can be found in Algorithm \ref{a:fine}, where `embed($G,d$)' on line \ref{a:embed} refers to using any multi-dimensional scaling method (such as cMDS, Laplacian Eigenmaps, etc.) to embed the dissimilarity matrix $G$ into a Euclidean space with dimension $d$.

\begin{algorithm}[t]
\caption{Fisher Information Non-parametric Embedding}
\label{a:fine}
    \begin{algorithmic}[1]
        \REQUIRE Collection of data sets $\mX=\{\vX_1,\vX_2,\ldots,\vX_N\}$; the desired embedding dimension $d$
        \FOR{$i=1$ to $N$}
            \STATE Calculate $\hat{p}_i(\vx)$, the density estimate of $\vX_i$
        \ENDFOR
        \STATE Calculate $G$, where $G(i,j)=\hat{D}_F(p_i,p_j)$, the geodesic approximation of the Fisher information distance
        \STATE $\vY=\textrm{embed}(G,d)$ \label{a:embed}
        \ENSURE $d$-dimensional embedding of $\mX$, into Euclidean space $\vY\in \Real^{d\times N}$
    \end{algorithmic}
\end{algorithm}

\subsection{Kernel Density Estimation}
\label{A:KDE}
The PDF of data set $\vX=\left[x_1,\ldots,x_n\right]$, $x_i\in\Real^m$, can be approximated with a kernel density estimate (KDE). This non-parametric method estimates density as the normalized sum of identical densities centered about each data point within the set:
\[
    \hat{p}(x)=\frac{1}{Nh}\sum_{i=1}^N{K\left(\frac{x-x_i}{h}\right)}
,\]
where $K$ is some kernel satisfying the properties
\[
K(x)\ge0
,\]
\[\int K(x)\,dx=1,\]
and $h$ is the bandwidth or smoothing parameter. We utilize a Gaussian distribution for our kernel,
\[
        K(x)=\frac{1}{(2\pi)^{(m/2)}|\Sigma|^{1/2}}\exp\left(-\frac{1}{2}x^T\Sigma^{-1}x\right)
    ,\]
where $\Sigma$ is the covariance matrix, as they have the quadratic properties that will be useful in implementation. The kernel parameters can be estimated using methods such as\cite{Silverman:86,Terrell:JASA90}.
\subsection{Orthonormality Constraint on Gradient Descent}
\label{A:Orth}
We derive the orthonormality constraint for our gradient descent optimization in the following manner; solving
\[A=\arg\min_{A:AA^T=I} J(A)
,\]
where $I$ is the identity matrix. Using Lagrangian multiplier $M$, this is equivalent to solving
\[A=\arg\min_A \tilde{J}(A)
,\]
where $\tilde{J}(A)=J(A)+{\rm tr}(A^TMA)$. We can iterate the projection matrix A, using gradient descent, as:
\Eq{\label{e:a_update}
A_{i+1}=A_i-\mu\frac{\partial}{\partial A}\tilde{J}(A_i)
,}
where $\frac{\partial}{\partial A}\tilde{J}(A)=\frac{\partial}{\partial A} J(A)+(M+M^T)A$ is the gradient of the cost function w.r.t. matrix $A$. To ease notation, let $\Delta \triangleq \frac{\partial}{\partial A} J(A_i)$ and $\tilde{\Delta} \triangleq \frac{\partial}{\partial A} \tilde{J}(A_i)$. Continuing with the constraint $A_{i+1}A_{i+1}^T=I$, we right-multiply (\ref{e:a_update}) by $A_{i+1}^T$ and obtain
\[
0=-\mu A_i\tilde{\Delta}^T-\mu\tilde{\Delta} A_i^T+\mu^2\tilde{\Delta}\tilde{\Delta}^T
,\]
\Eq{\label{e:derive1}
\mu\tilde{\Delta}\tilde{\Delta}^T=\tilde{\Delta}A^T+A \tilde{\Delta}^T
,}
\[
\mu(\Delta + (M + M^T)A)(\Delta + (M+M^T)A)^T = (\Delta A(M+M^T)A)A^T + A(\Delta A^T(M+M^T)A)
.\]
Let $Q=M+M^T$, hence $\tilde{\Delta}=\Delta+QA$. Substituting this into (\ref{e:derive1}) we obtain:
\[
\mu(\Delta \Delta^T+QA\Delta^T + \Delta A^TQ + QQ^T) = \Delta A^T + A\Delta^T + 2Q
.\]
Next we use the Taylor series expansion of $Q$ around $\mu=0$: $Q=\sum_{j=0}^\infty\mu^j Q_j$. By equating corresponding powers of $\mu$ (i.e.~$\frac{\partial^j}{\partial\mu^j}|_{\mu=0}=0$), we identify:
\[
Q_0=-\frac{1}{2}(\Delta A^T + A\Delta ^T)
,\]
\[
Q_1=\frac{1}{2}(\Delta + Q_0A)(\Delta + Q_0A)^T
.\]
Replacing the expansion of $Q$ in $\tilde{\Delta}=\Delta + QA$:
\[
\tilde{\Delta}=\Delta-\frac{1}{2}(\Delta A^T+A\Delta^T)A + \mu\, Q_1 A + \mu^2\, Q_2A +\ldots
.\]
Finally, we would like to assure a sufficiently small step size to control the error in forcing the constraint due to a finite Taylor series approximation of $Q$. Using the $L_2$ norm of $\tilde{\Delta}$ allows us to calculate an upper bound on the Taylor series expansion:
\[
\|\tilde{\Delta}\| \, \leq \, \|\Delta-\frac{1}{2}(\Delta A^T+A\Delta^T)A\| + \mu\,\| Q_1 A\| + \mu^2\,\| Q_2A\| +\ldots
.\]
We condition the norm of the first order term in the Taylor series approximation to be significantly smaller than the norm of the zeroth order term. If $\mu \ll \|\Delta-\frac{1}{2}(\Delta A^T+A\Delta^T)A\| / \|Q_1 A\|$ then:
\Eq{ \label{e:final_grad}
\frac{\partial}{\partial A}\tilde{J}(A)=\frac{\partial}{\partial A} J(A)-\frac{1}{2}\left((\frac{\partial}{\partial A} J(A)) A^T+A(\frac{\partial}{\partial A} J(A)^T)\right)A
}
is a good approximation of the gradient constrained to $A A^T=I$. We omit the higher order terms as we experimentally find that they are unnecessary, especially as even $\mu^2\rightarrow 0$. We note that while there are other methods for forcing the gradient to obey orthogonality \cite{Ruda&Herrera:LNCS06}, we find our method is straight-forward and sufficient for our purposes.



\bibliography{ref}
\bibliographystyle{IEEEbib}

\end{document}